\documentclass[10pt,journal,compsoc]{IEEEtran}

\ifCLASSOPTIONcompsoc
  \usepackage[nocompress]{cite}
\else
  \usepackage{cite}
\fi

\usepackage{amsmath,amsfonts}
\usepackage{algorithmic}
\usepackage{graphicx}
\usepackage{textcomp}
\usepackage[acronym]{glossaries}
\usepackage{soul}
\usepackage{subfigure}
\usepackage{booktabs}
\usepackage{multirow}
\usepackage{comment}
\usepackage{enumitem}
\usepackage{amsmath}
\usepackage{layouts}
\usepackage[ruled,vlined]{algorithm2e}
\newacronym{radar}{RADAR}{RAdio Detection And Ranging}
\newacronym{lidar}{LIDAR}{LIght Detection and Ranging}
\newacronym{lstm}{LSTM}{Long Short-Term Memory}
\newacronym{cnn}{CNN}{Convolutional Neural Network}
\newacronym{knn}{K-NN}{K-Nearest Neighbor}
\newacronym{mlp}{MLP}{Multi-Layer Perceptron}
\newacronym{tfn}{TFN}{Transformer Network}
\newacronym{mpnn}{MPNN}{Message Passing Neural Network}
\newacronym{svm}{SVM}{Support Vector Machine}
\newacronym{gflop}{GFLOP}{Giga Floating Point OPeration}
\newacronym{mb}{MB}{Mega Bytes}
\newacronym{ahc}{AHC}{Agglomerative Hierarchical Clustering}
\newacronym{iot}{IoT}{Internet of Things}
\newacronym{tfnet}{TFNet}{Transformer Network}
\newacronym{gpu}{GPU}{Graphical Processing Unit}
\newacronym{adc}{ADC}{Analog to Digital Conversion}
\newacronym{gnn}{GNN}{Graph Neural Network}
\newacronym{nlp}{NLP}{Natural Language Processing}
\newacronym{rnn}{RNN}{Recurrent Neural Network}
\newacronym{fmcw}{FMCW}{Frequency-Modulated Continuous Wave}
\newacronym{mimo}{MIMO}{Multiple-Input and Multiple-Output}
\newacronym{fft}{FFT}{Fast Fourier Transform}
\newacronym{cfar}{CFAR}{Constant False Alarm Rate}
\newacronym{ubpg}{UBPG}{Upper Body Point Cloud Gestures}
\newacronym{rf}{RF}{Radio Frequency}
\newacronym{dec}{DEC}{Dynamic Edge Convolution}
\newacronym{auc}{AUC}{Area Under ROC Curve}
\newacronym{rl}{RL}{Reinforcement Learning}
\newacronym{csi}{CSI}{Channel State Information}

\newcommand{\model}{Tesla}
\newcommand{\system}{Tesla-Rapture}
\newcommand{\lighmodel}{Tesla-V}

\newcommand{\layername}{TeslaConv}
\newcommand{\proposedknn}{Temporal \gls{knn}}
\DeclareMathOperator*\concat{\mathbin{\|}}

\newcommand{\easy}{\textsc{Easy}}
\newcommand{\complex}{\textsc{Complex}}
\newcommand{\all}{\textsc{All}}

\newcommand{\fakeparagraph}[1]{\vspace{1mm}\noindent\textbf{{#1.}}}

\begin{document}
\title{\system: A Lightweight Gesture Recognition System from mmWave Radar Point Clouds }

\author{Dariush~Salami\IEEEauthorrefmark{1},
        Ramin~Hasibi\IEEEauthorrefmark{1},
        Sameera~Palipana,~Petar~Popovski, Tom~Michoel,~and~Stephan Sigg
\IEEEcompsocitemizethanks{

\IEEEcompsocthanksitem D. Salami, S. Palipana, and S. Sigg are with the Department of Communications and Networking, Aalto University, Espoo,
Finland.\protect\\
E-mails: \{dariush.salami, sameera.palipana, stephan.sigg\}@aalto.fi

\IEEEcompsocthanksitem R. Hasibi and T. Michoel are with the Department of Informatics, University of Bergen, Bergen, Norway.\protect\\
E-mails: \{ramin.hasibi, tom.michoel\}@uib.no

\IEEEcompsocthanksitem P. Popovski is with the Department of Electronic Systems, Aalborg University, Aalborg, Denmark.\protect\\
E-mail: petarp@es.aau.dk
\protect\\
\protect\\
\IEEEauthorrefmark{1} Both authors contributed equally to this research.
}
}

\markboth{Journal of \LaTeX\ Class Files,~Vol.~14, No.~8, August~2015}%
{Shell \MakeLowercase{\textit{et al.}}: Bare Advanced Demo of IEEEtran.cls for IEEE Computer Society Journals}
\IEEEtitleabstractindextext{%
\begin{abstract}
We present \system, a gesture recognition interface for point clouds generated by mmWave Radars. 
State of the art gesture recognition models are either too resource consuming or not sufficiently accurate for integration into real-life scenarios using wearable or constrained equipment such as IoT devices (e.g. Raspberry PI), XR hardware (e.g. HoloLens), or smart-phones. 
To tackle this issue, we developed \model, a Message Passing Neural Network (MPNN) graph convolution approach for mmWave radar point clouds. The model outperforms the state of the art on two datasets in terms of accuracy while reducing the computational complexity and, hence, the execution time. 
In particular, the approach, is able to predict a gesture almost 8 times faster than the most accurate competitor.
Our performance evaluation in different scenarios (environments, angles, distances) shows that \model~generalizes well and improves the accuracy up to 20\% in challenging scenarios like a 
through-wall setting and sensing at extreme angles.
Utilizing \model, we develop \system, a real-time implementation using a mmWave Radar on a Raspberry PI 4 and evaluate its accuracy and time-complexity.
We also publish the source code, the trained models, and the implementation of the model for embedded devices. 
\end{abstract}

}

\maketitle

\IEEEdisplaynontitleabstractindextext

\IEEEpeerreviewmaketitle

\IEEEraisesectionheading{\section{Introduction}\label{sec:introduction}}
\begin{figure*}
    \centering
    \includegraphics[width=\linewidth]{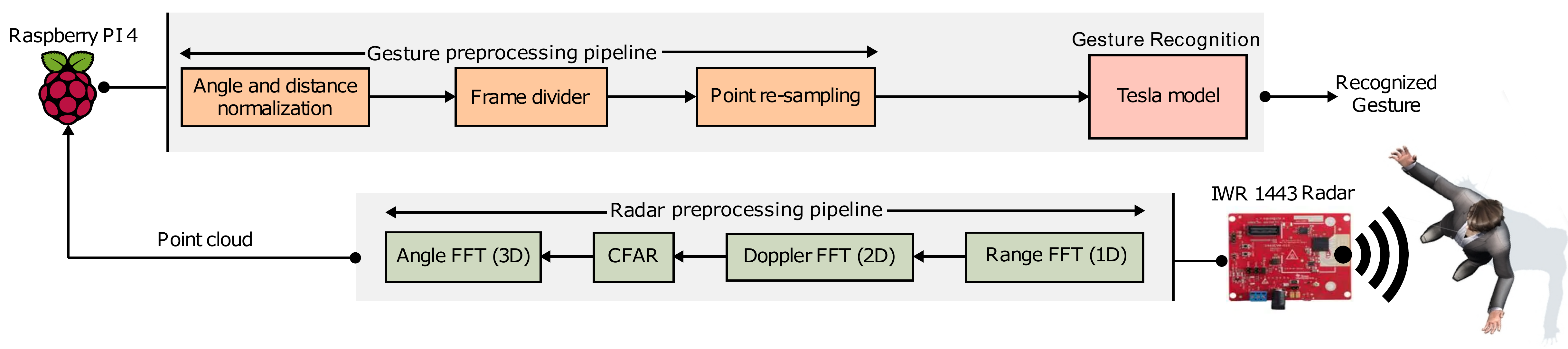}
    \caption{Overview structure of \system. The  radar transforms the IQ samples into a point cloud through the radar processing pipeline and this is fed to a Raspberry Pi 4 for further processing and infer the gestures.}
    \label{fig:tesla_rapture}
\end{figure*}
Gesture recognition is a substantial part of human-computer interaction systems in domains such as smart homes~\cite{wan2014gesture}, vehicular applications~\cite{ohn2014hand}, and human-robot interaction~\cite{liu2018gesture}. To do so, movement is captured (e.g. RGB, depth, ultrasound, Radar, etc.), pre-processed, and the data is finally fed to a classification model to recognize gestures and trigger control commands in the system.

Traditional work utilizes ultrasound~\cite{kalgaonkar2009one, przybyla201412}, wearable sensors~\cite{lu2014hand, zhang2015tomo}, or cameras~\cite{gupta2016online, pisharady2015recent} as gesture sensors. 
However, they have drawbacks that can limit real-world deployment, such as limited sensing range, discomfort of wearing, or the risk of privacy leakage. 
Electromagnetic radiation, another alternative solution for mid-air, device-free gesture recognition, facilitates a variety of interaction modalities such as WiFi, radar, infra-red, or RGB-depth sensors. 
Even though recently, gesture recognition using Wi-Fi \gls{csi} has been popular, it cannot recognize fine-grained gestures  due to the limitations imposed by the wavelength.
Radar sensing has some common features to WiFi in that it is robust to weather conditions, does not require lighting, and can penetrate thin, non-metallic surfaces (depending on the wavelength). Additionally, it can operate in mono-static configuration, providing 3D spatial information through \gls{mimo} capabilities.
In addition, their high millimeter wave operating frequencies allow for small form factors, so that the sensor can be mounted on miniature devices and provides fine-grained gesture recognition   through  large antenna arrays. 
Millimeter waves are non-ionizing and thus not dangerous to the human body.

Many real-life scenarios that involve gesture recognition require computationally tractable models that can be implemented on off-the-shelf processing units to provide real-time detection functionality as well as reasonable accuracy. 
Traditional recognition often extracts hand-crafted features from the data and feeds them to a classification algorithm such as \gls{svm}, Naïve Bayes or decision tree \cite{yun2009automatic, huang2009vision}.
With the advent of deep learning as an automatic feature learning approach, gesture recognition models have shown a significant improvement in accuracy~\cite{molchanov2016online, yang2018making}. 
On the downside, the need for feeding the entire data to the feature extraction pipeline results in computationally expensive models. 
Consequently, most deep learning based models cannot be directly implemented on constrained devices to provide real-time user experience as they require high processing power. 

The input representation plays an important role in both accuracy and time-complexity of deep learning based systems. 
RGB images \cite{yang2018making,molchanov2016online, abavisani2019improving}, depth images\cite{yang2018making,molchanov2016online, abavisani2019improving}, spectrograms of Doppler signals \cite{kim2016hand}, and point clouds\cite{qi2017pointnet,min2020pointlstm, palip2021pantomime, salami2020motion} are commonly used representations. 
Among these, point clouds, i.e., unordered sets of points in space, are the standard output of a wide range of sensors~\cite{qi2017pointnet}. 
Furthermore, converting the raw \gls{adc} data from the antenna arrays to point clouds massively reduces the data size by several magnitudes (e.g. GBytes to MBytes), resulting in faster data transfer, pre-processing, and inference time. 
Unlike spectrograms of Doppler signals, point clouds are easily interpretable since the motions occur in a 3D space.

Point cloud processing models are categorized into multi-view, volumetric, and direct point cloud interpretation.
Multi-view techniques~\cite{su2015multi, yu2018multi} project the input point cloud onto 2D-planes for 2D image processing, making predictions according to the fused latent features.
Volumetric techniques~\cite{maturana2015voxnet, riegler2017octnet} produce voxels in 3D space (equivalent to pixels in 2D) from input point clouds and extract features through 3D volume processing.
Direct point cloud processing~\cite{qi2017pointnet,wang2019dynamic} extracts features from the input point cloud without intermediate representation. 
The latter approach should guarantee permutation invariance w.r.t. points to effectively cope with $n!$ permutations of a point cloud with $n$ points. 
Since processing in multi-view and volumetric techniques is lossy, computationally intensive and thus time consuming, direct processing of point clouds is the most promising in terms of accuracy and run-time.

To solve the model complexity problem, we propose a direct point cloud processing method, \model~(\textbf{TE}mporal graph \textbf{S}e\textbf{L}f \textbf{A}ttention convolution), a \gls{mpnn} graph convolution based architecture tailored to sparse point clouds generated by mmWave radars. 
Utilizing the unique properties of mmWave radar point clouds, we introduce a novel \proposedknn~algorithm to dynamically model the temporal evolution of the point cloud over successive frames as a graph structure, and a novel self-attention \gls{mpnn} based graph convolution layer called \layername~to process the generated graph and infer the gestures. 
Unlike \gls{rnn} based models, which iteratively fuse spatial features of each time frame, our method takes advantage of a novel graph convolution with a single forward pass to capture the temporal evolution. As a result, this approach outperforms the state of the art in terms of accuracy and computational complexity, which positions it for embedded devices and real-time settings. In particular, \model~is ahead of state of the art by a margin of up to 4.2\% and 2.9\% on main settings as well as 21\% in challenging scenarios of two different datasets. Moreover, the model is 8 times faster and has almost 40 times less computational complexity  than the most accurate competitor when it comes to inference time and \glspl{gflop} respectively.

Given the widespread usage of Raspberry PI in \gls{iot} world from human-robot interaction \cite{odoemelem2020low, soni2018artificial} to smart-home applications \cite{vujovic2015raspberry, jain2014raspberry}, we integrate the proposed model in a system called \system~(\textbf{\model} for \textbf{RA}dar generated \textbf{P}oint cloud ges\textbf{TURE}) on Raspberry PI 4, the architecture of which is depicted in Fig. \ref{fig:tesla_rapture}.

Our main contributions are:
\begin{itemize}[noitemsep,topsep=5pt]
    \item \proposedknn, a novel \gls{knn} algorithm to model the time dimension of point clouds as a temporal graph.
    \item To the best of our knowledge, we are the first to process motion point clouds using a graph convolution approach and develop a self-attention \gls{mpnn} to process the temporal graph built through the \proposedknn.
    \item A thorough performance evaluation on two datasets with different settings including diverse environments, distances, angles and speeds.
    \item An implementation on a Raspberry PI 4 in a real-time setting.
    \item A publicly available code, trained models, and Raspberry PI implementation for verification and follow-up research purposes.
\end{itemize}

\section{Related Work}
\label{sec:related_work}

\subsection{Gesture Recognition}
RGB cameras, RGB depth sensors, Leap Motion, mmWave radars, and WiFi are prominently mentioned in the literature for mid-air gesture recognition. 
Extensive surveys on vision-based gesture sensing were published by Wachs et al.~\cite{Wachs11} and Rautaray et al.~\cite{Rautaray12}.
These systems (e.g. MS Kinect) employ an RGB camera and an infrared depth sensor providing either 2D color frames, full-body 3D skeleton, or 3D point clouds~\cite{lun2015survey}. 
However, they are limited in darkness and occlusion, and the camera raises privacy concerns~\cite{Caine12_privacy}.

\gls{rf} gesture recognition can be distinguished into sub-6\,GHz and millimeter waves. 
The former leverages received signal strength~\cite{abdelnasser2015wigest} from commodity narrow-band devices, \gls{csi} from WiFi~\cite{li2016wifinger, ma2018signfi, venkatnarayan2018multi, virmani2017position}, Doppler~\cite{pu2013whole}, 
or radar~\cite{li2019making, zhao2019through}. 
However, the gesture recognition accuracy below 6~GHz is limited by its small bandwidth and a wavelength above 5~cm, so that antenna array apertures become too large. 
In contrast, mmWave sensing features high bandwidth (4-7~GHz) and antenna apertures of few centimeters.
For mmWave radars, gesture recognition is either model~\cite{lien2016soli, wei2015mtrack}
or data-driven~\cite{Berenguer2019GestureVLAD, panneer2020mmASL, singh2019radhar, wang2016interacting, zhao2019mid, meng2020gait}. 
Most data-driven approaches combine \gls{cnn} and \gls{rnn} modules to process Doppler, range-Doppler, and/or angle-Doppler features~\cite{Berenguer2019GestureVLAD, panneer2020mmASL, wang2016interacting}.
Since these features are dependent on relative direction of movement and angle granularity, complex tasks, such as distinguishing simultaneous movement of different body parts becomes challenging. 

\subsection{Static Point Clouds}
Point clouds are of different granularity depending on the modality used  to capture them.
Point clouds extracted from RGB-depth images and LiDAR are dense, while mmWave radars produce sparser point clouds~\cite{qian20203d} that do not highlight  the human skeletal structure~\cite{palip2021pantomime}. 
Recent years have witnessed the emergence of mmWave radar point cloud human sensing due to the availability of commercial hardware that is miniature and low cost (e.g. hand tracking~\cite{dong2020model}, gesture recognition~\cite{liu2020real, palip2021pantomime} activity recognition~\cite{singh2019radhar}, gait recognition~\cite{meng2020gait}, or positioning~\cite{zhao2019mid}).

In~\cite{qi2017pointnet}, \textit{PointNet} was introduced as the pioneering model for direct processing of 3D point clouds by extracting the features on a point-by-point basis and aggregating the features using a permutation-invariant pooling operation. 
In \textit{PointNet++}~\cite{qi2017pointnet2}, set abstraction modules for sampling and grouping neighbouring points in each processing layer has been added to better represent spatial features.

At the same time, by applying \glspl{cnn} on graphs, graph convolution approaches have emerged~\cite{Kipf:2016tc}. 
Modeling point clouds as graphs in which nodes correspond to points and edges connect points to their nearest neighbours in Euclidean space, makes it possible to apply graph convolution principles. 
In particular, a message passing algorithm, known as \gls{mpnn}, is utilized to gradually propagate each point's features as a message to its neighbours and to aggregate the incoming messages with the features of the point itself~\cite{2017Gimessagepassing}. Based on \gls{mpnn},~\cite{wang2019edgeconv} introduced \textit{\gls{dec}} by redefining the graph using \gls{knn} at each convolution layer and the messages as Euclidean distance between neighbouring points. Although \gls{dec} performs well on the shape classification task, it fails to capture temporal dependency in mid-air gesture recognition. This shortcoming specifically affects the gestures with similar aggregated point clouds through time dimension, e.g., swipe-left and swipe-right gestures. To address this issue, we reflect the temporal evolution of gestures in graph structures, i.e, each point can only connect to the points from previous frames.

\subsection{Dynamic Point Clouds}

Previous attempts at capturing spatio-temporal features of dynamic point clouds include using a combination of \gls{rnn} with either 3D\gls{cnn} or PointNet layers~\cite{salami2020motion, owoyemi2018spatiotemporal, palip2021pantomime}, as well as using a modified \gls{rnn} layer to propagate information temporally while preserving the spatial structure in each frame~\cite{min2020pointlstm}. 
In real-world applications, these models are constrained by their high computational complexity and restricted generalizability on point clouds generated in different settings. However, given the sparsity of the mmWave radar point clouds in each frame (in average 5-10 points per frame), extraction of frame-wise spatial features does not contribute to the latent representation of gestures. Moreover, the recurrent pipeline of \gls{rnn}-based model increases the computational complexity. To tackle this problem, we capture the temporal dependency reflected in the graph structure using a single pass of the proposed \gls{mpnn} model. To further increase the performance of the model, we integrate the self-attention mechanism~\cite{vaswani2017attention} to increase the impact of important parts of the input data while fading out the rest. 


 

\section{ Point Clouds from mmWave Radars}
We use point cloud datasets from the Texas Instruments IWR1443\footnote{https://www.ti.com/product/IWR1443} sensor, a \gls{fmcw}–\gls{mimo} radar sensor that operates in the 77 GHz RF band. A radar transmitter antenna (Tx) emits an electromagnetic signal, which is reflected and scattered by objects in the environment, before it is captured again by a receiving antenna (Rx). An \gls{fmcw} signal is used for range estimation of the reflecting objects and a \gls{mimo} configuration is utilized to compute both elevation and azimuth angles \cite{palip2021pantomime}. A coordinate transformation of the range, azimuth and elevation angles of the detected objects yields the point cloud in a $x$-$y$-$z$ coordinate system.
The following signal processing pipeline achieves the point cloud representation from \gls{adc} data.
\subsection{Point Cloud Generation}
\label{sec:point_cloud_generation}
\begin{figure}
    \centering
\includegraphics[width=\columnwidth]{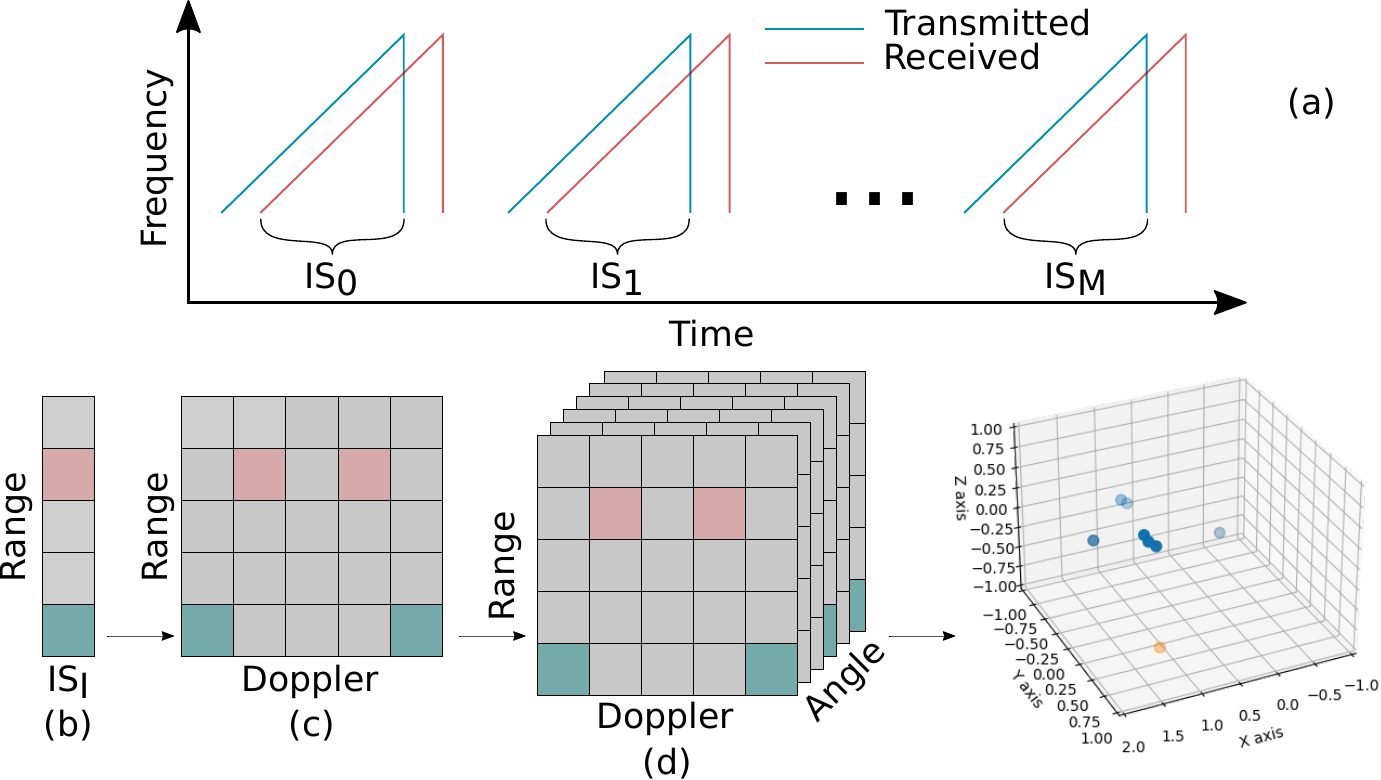}
    \caption{(a) The transmitted and the reflected chirps are shown in the frequency domain. (b) The range of the detected objects after applying 1D-\gls{fft} on the intermediate signal. (c) The velocity of the detected objects after 2D-\gls{fft}. (d) The angle of the detected objects after applying 3D-\gls{fft} on the data from multiple antennas}
    \label{fig:fmcw_piepline}
\end{figure}
The processing unit on the evaluation kit of the radar applies a four step preprocessing pipeline to obtain point clouds.

\textit{Range-\gls{fft} (1D)}: The radar sends a chirp signal (Fig. \ref{fig:fmcw_piepline}.a), i.e, a signal with linearly increasing carrier frequency, and produces an intermediate frequency signal by mixing the transmitted and received chirps and low pass filtering. 
The distance to the reflecting object is proportional to the intermediate frequency, which is computed using the \gls{fft} operation on the mixed signal (Fig. \ref{fig:fmcw_piepline}.b).
    
\textit{Doppler-\gls{fft} (2D)}: Two or more time-separated chirps are required to estimate the radial velocity of an object. 
The phase difference between two chirps at the range-\gls{fft} peak is proportional to the radial velocity of the detected object (2D-\gls{fft} or Doppler-\gls{fft}) which is shown in Fig. \ref{fig:fmcw_piepline}.c. 
    
\textit{\gls{cfar}}: The \gls{cfar} detection algorithm~\cite{richards2010principles} is used to separate reflecting objects from noise. The summation of the Doppler-\gls{fft} matrices creates a pre-detection matrix.
The \gls{cfar} algorithm identifies peaks in the pre-detection matrix that correspond to the detected objects. The elements with gray color in Fig. \ref{fig:fmcw_piepline} show the noisy points that are filtered by \gls{cfar} algorithm.
    
\textit{Angle-\gls{fft} (3D)}: For each object in the \gls{cfar} algorithm, an \gls{fft} of the angle is performed on the corresponding \gls{cfar} peaks across multiple Doppler-\gls{fft}s (Fig. \ref{fig:fmcw_piepline}.d). 
Velocity-induced phase changes are Doppler-corrected before computing the angle-\gls{fft}.

\subsection{Point Cloud Properties}
\label{sec:point_cloud_properties}
The point cloud generated from the above process has unique spatial and temporal properties.

\subsubsection{Spatial properties} The point cloud is sparse and the skeleton structure of the human is not apparent in individual frames. 
The radar captures more points during motion than during stationary phases of an object or subject. 
This is attributed to the signal processing tool chain used for the radar. First, the point cloud is extracted through range-\gls{fft}, 
Doppler-\gls{fft}, \gls{cfar}, and angle-\gls{fft} operations as described above. 
The \gls{cfar} algorithm relies on range and Doppler dimensions to detect an object, so that the detected cloud points are triggered due to the motion and intensity of the reflection. This property is used to filter stationary reflections in the environment.

The gestures in the horizontal plane have a higher granularity than in the vertical plane, since the radar has more antenna elements in the azimuth direction. 
Eight virtual elements can resolve an angle of $14.3^\circ$, in contrast to only $57^\circ$ via two virtual antennas in the elevation direction. 
Another reason is the sensitivity of the \gls{cfar} algorithm in the Doppler direction. 

For reflecting objects or subjects close to the sensor, the larger radar cross-section results in denser point clouds. 
For instance, representations of arms or hands become less sparse in short distance case. 
Additionally, for reflections off objects in a distance $D$, the spacing between points captured at a resolution of $\theta$ is proportional to $D\cdot\theta$. This causes the point cloud to have a distance dependant density and causes the trained model accuracy to deteriorate with increasing distance. 

\subsubsection{Temporal properties}
The \gls{cfar} algorithm collapses points that are detected over a specific fixed temporal duration $t_\Delta$ into frames. 
The number of cloud points is variable across frames. 
Even though the skeletal structure of the body is not apparent in individual frames, an arm's motion constructs a spatio-temporal structure in the direction of motion over successive frames.
These unique spatio-temporal structures in the point cloud for different gestures can be exploited for motion gesture recognition.

\subsection{Comparison with RGB-D Point Clouds}
\label{sec:comparison_with_rgb_d_point_cloud}
\begin{figure}
    \centering
\includegraphics[width=\columnwidth]{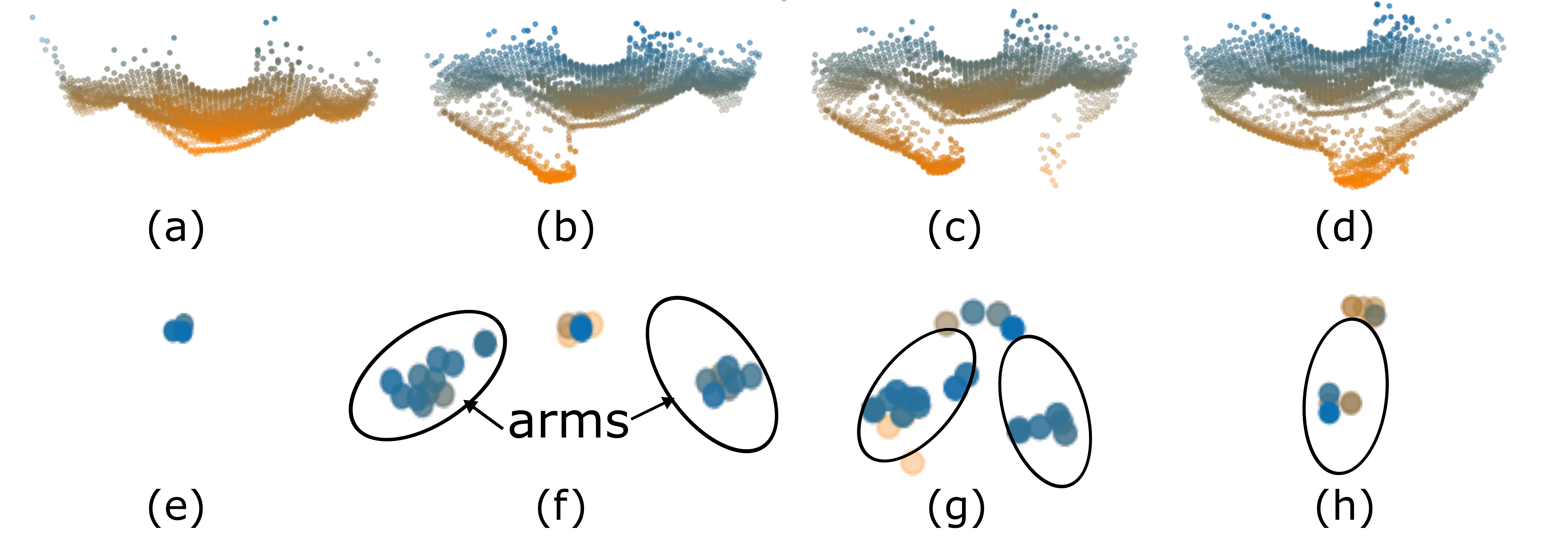}
    \caption{(a), (b), (c), and (d) are point clouds of four frames from a single gesture of the \gls{ubpg} dataset which is essentially a closing in of the two arms from a wider position. A similar gesture captured by the mmWave radar is shown in (e), (f), (g), and (h) over four frames.}
    \label{fig:rgbd_comp}
\end{figure}

Compared to RGB-D point clouds, mmWave point clouds are sparse. 
We illustrate this in Fig.~\ref{fig:rgbd_comp} using similar gestures from two datasets. 
In particular, we utilize the \gls{ubpg} RGB-D gesture point cloud dataset~\cite{owoyemi2018spatiotemporal}, and the Pantomime dataset~\cite{palip2021pantomime} (point clouds of gestures captured by a mmWave radar). 
Indeed, mmWave point clouds hold little information in each frame. 
Still, stretched over four frames, a motion gesture is evolving, that describes the two clusters of points corresponding to the arms to close in. 
Specifically, the spatial relation between points in each individual frame is less expressive to infer a gesture than the temporal dependencies of points across consecutive frames.

\section{Proposed Model}
\begin{figure*}[!ht]%
    \centering
    \includegraphics[width=\linewidth]{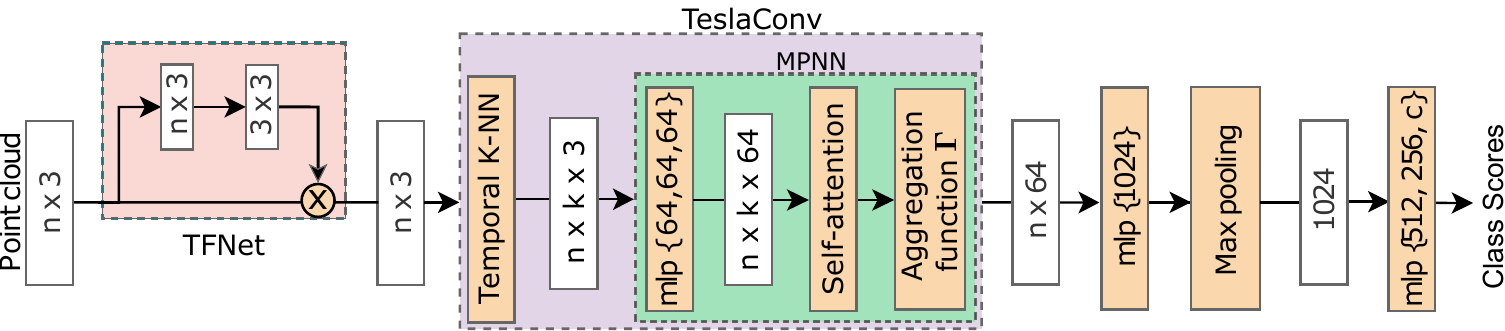}
    \caption{The architecture of \model- Having multiplied the input point cloud by a $3\times3$ spatial transformation matrix of \gls{tfnet}, the transformed output is fed into \layername. In \layername~a temporal graph is created using \proposedknn~module and the proposed message passing scheme (section \ref{sec:mpnn}) is applied. Afterwards, to represent the gesture as a fixed-sized vector, an \gls{mlp} of size 1024 followed by a max pooling is performed. Finally, a three layered \gls{mlp} with respective sizes of 512, 256, and $c$ (the number of classes) is used to predict the class scores of the gesture.}
    \label{fig:sys_archi}
\end{figure*}

In this section, we describe \model, an \gls{mpnn} based graph convolution approach tailored for inferring gestures from motion point clouds. 

The architecture of \model~is depicted in Fig. \ref{fig:sys_archi}. 
First, in order to make the prediction model robust against possible spatial transformations of input gestures (e.g., rotation, translation, scaling, etc.), we apply a \gls{tfnet}~\cite{jaderberg2015transform} module on the input point cloud. This trainable module is responsible for producing a dynamic transformation for each input gesture's entire feature map to transform the possibly skewed points to a rigid, uniform, and canonical point cloud, which in turn makes the recognition in the following layers simple. Next, we apply our proposed \layername~layer on the output of \gls{tfnet}, which includes two steps: \textit{Graph Generation} and \textit{Graph Processing}. In the Graph Generation phase, a temporal graph is created from motion point clouds through the proposed \proposedknn~algorithm, which connects each point to its nearest neighbors from previous frames to reflect the temporal pattern of gesture. In the Graph Processing step, we apply the proposed \gls{mpnn} scheme that learns the representation of each point according to the structure of the generated graph. Additionally, we optimize this layer by integrating an self attention mechanism in the message passing scheme to improve the performance of the graph processing. Furthermore, it decreases the computational complexity of the model by eliminating the need for removing outliers of the dataset explicitly.
In the following we will present more details about each step of the \layername.

\subsection{Graph Generation}
\label{sec:graph_generation}

Consider a point cloud $X=\{x_1, ..., x_n\} \subseteq \mathbb{R}^F$ where each point is represented by a feature set of $x_i=\{f_i^1, ..., f_i^F\}$. In motion point clouds the frame number $f_i^s$ of each point is also a dimension of the feature set, i.e., $f_i^s\in x_i$. The \gls{knn} graph $\mathcal{G}=\{X, \mathcal{E}\}$ is obtained through the \gls{knn} algorithm where $\mathcal{E} \subseteq X \times X$ is the set of directed edges between each point and its closest neighbours in the Euclidean space.

\begin{figure*}
    \centering
    \includegraphics[width=\linewidth]{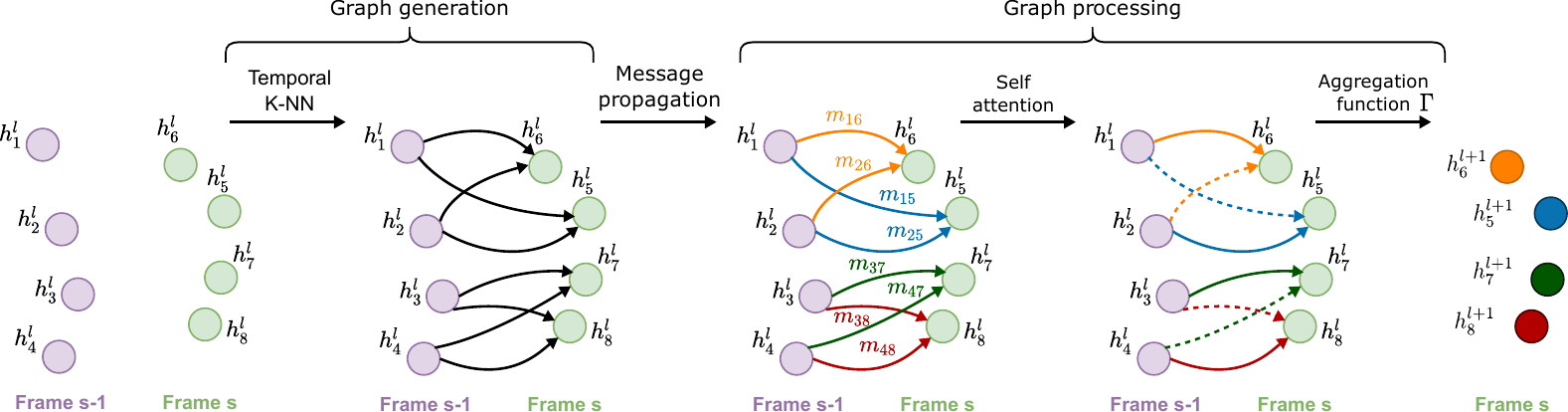}
    \caption{\layername~layer- For points in each frame, a directed edge is connected from the nearest neighbours in the previous frames through \proposedknn. Next, messages are propagated according to the direction of the edges and a multi-head self-attention is performed on them. Finally, the representation of each point is obtained by $\Gamma$ aggregation function on incoming messages.}
    \label{fig:layer_arch}
\end{figure*}

\begin{figure*}
    \centering
        \subfigure[]{
        \raisebox{0.3\height}{\includegraphics[width=.22\textwidth]{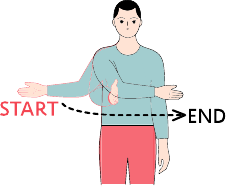}
        }}
    \hfill
    \subfigure[]{
        \includegraphics[width=.28\textwidth]{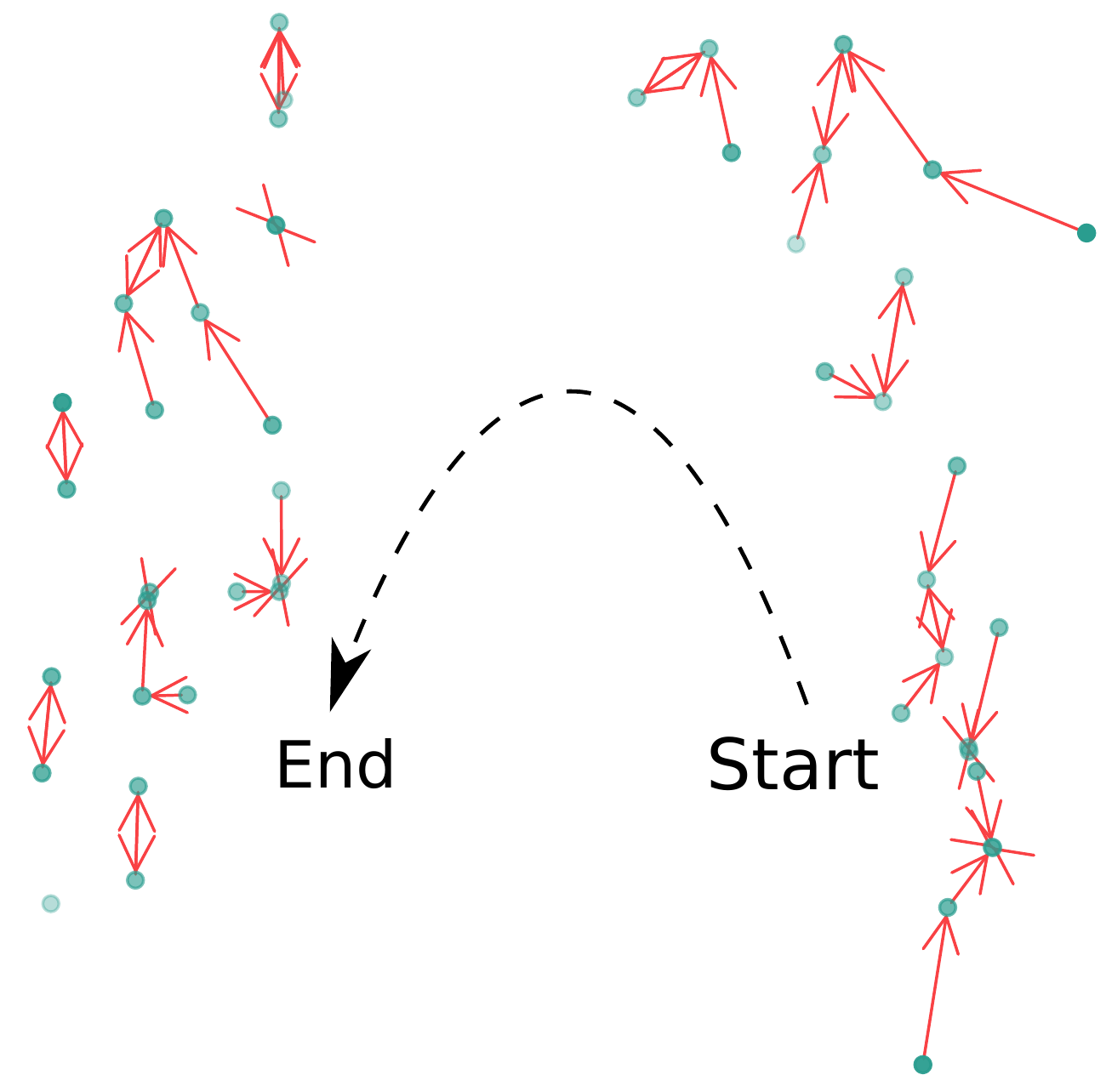}%
        }
    \hfill
    \subfigure[]{
        \includegraphics[width=.28
        \textwidth]{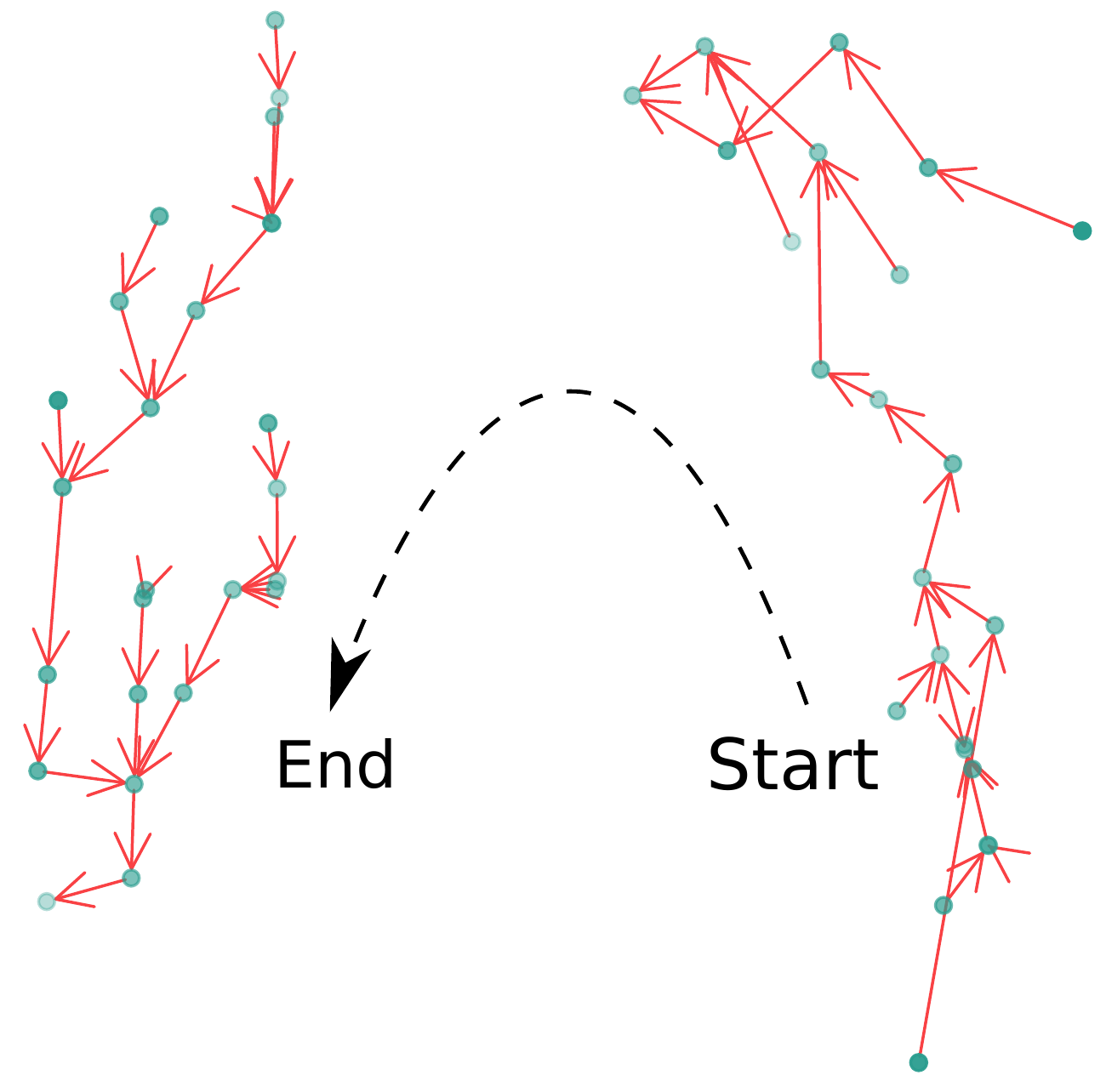}%
        }
    
    \caption{Intuition behind \proposedknn- (a) The schematic of the swipe-left gesture from the Pantomime dataset (b) Generated graph using \gls{knn} (c) Generated graph using \proposedknn~ with $\alpha=100$. Both point clouds are shown from a top view and K is equal to 1 for simplicity.}
    \label{fig:knn_intution}
\end{figure*}

As illustrated in Fig.~\ref{fig:layer_arch}, in the graph generation phase, for each point, we use \proposedknn~to find the nearest neighbors only from the previous frames. For swipe-left gesture in Pantomime dataset, the comparison between \gls{knn} and \proposedknn~in the graph structure is shown in Fig. \ref{fig:knn_intution}. The trend in the direction of the arrows in Fig. \ref{fig:knn_intution}.(c) shows the temporal evolution of the gesture whereas that of Fig. \ref{fig:knn_intution}.(b) is irrelevant to the temporal pattern.

In the first step of \proposedknn, we normalize the feature set of each input point using batch-wise min-max normalization.
\begin{equation}
    x_i = \frac{x_i - min(x)}{max(x)-min(x)}.
    \label{eq:knn_norm}
\end{equation}
in which, $min(x)$ and $max(x)$ are the minimum and maximum values of each dimension of $x$ over a batch of input gestures, respectively.
In the second step, we multiply the temporal dimension of $x_i$ ($f_i^s$) by a hyperparameter $\alpha$ to control the trade-off between temporal and spatial features. Setting $\alpha$ to a large number (e.g., 100) forces the model to find the nearest neighbors only from the previous frame, while small numbers of $\alpha$ (e.g., 0) gives the model more freedom in choosing the nearest neighbors from the whole non-masked set.

\begin{equation}
    f_i^s = \alpha f_i^s.
    \label{eq:alpha_normalization}
\end{equation}

To find the nearest neighbors only from previous frames, we introduce a masking scheme. The masked set of points $\mathcal{F}_{x_i}$ for $x_i$ is obtained through:

\begin{equation}
    \mathcal{F}_{x_i} =
    \{x_j : \forall x_j \in X, f_j^s > f_i^s\}
    \label{eq:knn_points_from_frame}
\end{equation}

Furthermore, the distance between two points is defined as the Euclidean distance of all the corresponding features of points including $f_i^s$ and is calculated according to:

\begin{equation}
    D_{x_i,x_j} =
    \begin{cases}
    ||x_i - x_j|| : x_i,x_j \in X ,& \text{if } x_j \notin \mathcal{F}_{x_i},\\
    \infty ,              & \text{otherwise},
    \end{cases}
    \label{eq:knn_points_from_frame}
\end{equation}
where $D_{x_i,x_j}$ denotes the distance between $x_i$ and $x_j$ and $||.||$ is Euclidean norm operator. Finally, the introduced masked distance function is used to find the nearest neighbors in \proposedknn.

\subsection{Graph Processing}
\label{sec:mpnn}
As shown in Fig. \ref{fig:layer_arch}, in the graph processing phase, the representation of each point is calculated through the proposed \gls{mpnn} layer based on the temporal graph. In each layer, the hidden representation of each point is updated through an aggregation function on the point features except for $f_i^s$ from the previous layer and the messages of its neighbours according to:

\begin{equation}
    \begin{aligned}
        h_{i}^0 &= x_i \setminus \{f_i^s\},\\
        h_{i}^l &=\underset{j:(i,j)\in \mathcal{E}}{\Gamma}M_\theta(h^{l-1}_i,h^{l-1}_j)),
    \end{aligned}
    \label{eq:dynamic_edge_convolution}
\end{equation}

in which, $h_i^l$ is the hidden representation of point $i$ in \gls{mpnn} layer $l$, $\setminus$ is the set subtraction operator, message function $M_{\theta}:\mathbb{R}^F \times \mathbb{R}^F \rightarrow \mathbb{R}^{F'}$ is a non-linear function with a set of trainable parameters $\theta$ and is usually implemented using \gls{mlp} architectures, $\Gamma$ is a channel-wise symmetric aggregation function (e.g. $\Sigma$, max, or mean) applied on the messages of the edge emanating from each neighbor. 

The choice of $M$ and $\Gamma$ significantly affects the properties and the performance of the model in Eq. \ref{eq:dynamic_edge_convolution}. For example, setting $M_\theta(h_i, h_j)=M_\theta(h_i)$ causes the model to only capture the global features of point clouds without considering the local structures. On the other hand, setting $M_\theta(h_i, h_j)=\bar{M}_\theta(h_i, h_j-h_i)$, provides information about the local relations of the neighbouring points. In this paper, we use the second setting of message function to help capture the local dependencies as well as the global structure.
\begin{figure}
    \centering
    \includegraphics{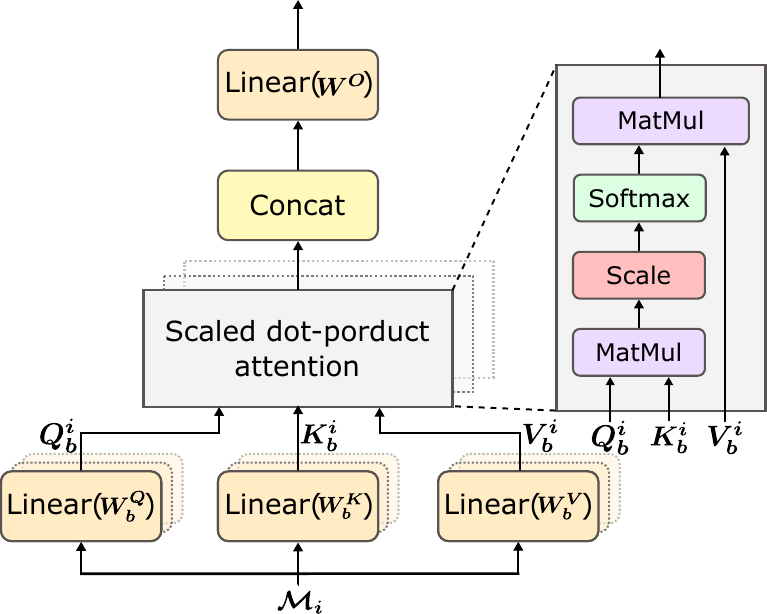}
    \caption{Multi-head Self Attention mechanism- \textbf{Linear} refers to multiplication with corresponding learnable weights (Eq. (\ref{eq:qkv_calculation}) \& Eq. (\ref{eq:multihead_message})), \textbf{Scaled dot-product attention} is formulated in Eq. (\ref{eq:scaled_dot_product}), and finally, \textbf{Concat} is the concatenation operation in Eq. (\ref{eq:multihead_message}).}
    \label{fig:self_attention}
\end{figure}

To decrease the effect of noisy points, we integrate a \textit{scaled-dot multi-head self-attention} mechanism \cite{vas2017attention} shown in Fig. \ref{fig:self_attention} into the message function. The goal is to let the incident edges to point $i$ decide their relative importance in determining the updated representation of the point. Let $\mathcal{M}_i = \bigcup_{j:(i,j)\in \mathcal{E}}\bar{M}_\theta(h_i, h_j-h_i)$ denote the set of the messages of incident edges for each point $i$. A set of query $Q_b^i$, key $K_b^i$, and value $V_b^i$ for point $i$ in a single-head self attention is calculated through:

\begin{equation}
    Q_b^i = \mathcal{M}_iW_b^{Q},~~
    K_b^i = \mathcal{M}_iW_b^{K},~~
    V_b^i = \mathcal{M}_iW_b^{V}
\label{eq:qkv_calculation}
\end{equation}%
where, $W_b^{Q}, W_b^{K}, W_b^{V}$ are trainable weights and $b$ is the head index.
Then, the single-head self attention on the messages of each point is calculated through:
\begin{equation}
    H_b^i(Q_b^i, K_b^i, V_b^i) = softmax\left(\frac{Q_b^i \times K_b^i}{\sqrt{|K_b^i|}}\right) \times V_b^i.
\label{eq:scaled_dot_product}
\end{equation}
in which, $\times$ is matrix multiplication operator.
Moreover, employing the multi-head approach allows the model to calculate the attention scores using different sub-spaces at the incident edges' messages as well as a more stable learning process. In this work we employ $m=8$ parallel attention layers with $f_o/m$ dimensions, where $f_o$ is the number of dimensions of incident messages after performing message function. The final multi-head output is obtained by
\begin{equation}
    A(\mathcal{M}_i) = (\concat_{b=1}^{m}H_b(Q_b^i, K_b^i,V_b^i))W^O,
\label{eq:multihead_message}
\end{equation}
Where $W^O$ are trainable weights and $\concat$ is the concatenation operator.
Thus, the message passing part of the \layername~layer (Eq. (\ref{eq:dynamic_edge_convolution})) can be updated as: 
\begin{equation}
\begin{aligned}
h_i^l = \underset{j:(i,j)\in \mathcal{E}}{\Gamma} A(\mathcal{M}_i^{l-1})
\end{aligned}
\label{eq:proposed_formula}
\end{equation}
For results on the effect of self-attention mechanism see section \ref{sec:hypertuning}.

\subsection{Permutation Invariance}
Since the permutation of points in $X$ does not alter the nature of the gesture, the prediction model should be permutation invariant with respect to the order of the input points. This can be proved in two steps for our approach.

First, \proposedknn, introduced in section \ref{sec:graph_generation}, uses symmetric aggregations ($min$ and $max$) and calculates the Euclidean distance between points which leads to a permutation invariant graph generation process.

Second, in this work we use $max$ as the aggregation function in Eq. (\ref{eq:proposed_formula}):
\begin{equation}
\begin{aligned}
h_i^l = \underset{j:(i,j)\in \mathcal{E}}{max} A(\mathcal{M}_i^{l-1})
\end{aligned}
\label{eq:proposed_formula_with_agg}
\end{equation}

Since $max$ in Eq. (\ref{eq:proposed_formula_with_agg}) and the global max-pooling function shown in Fig. \ref{fig:sys_archi} are symmetric functions, the output of the layer is permutation invariant w.r.t the input.

\section{Implementation}
In this section, we present the implementation details of \system~in terms of preprocessing pipeline, training and inference phases, and real-time gesture recognition interface on Raspberry PI 4.

\subsection{Preprocessing}
To prepare the data for \model~model and study the effect of different hyperparameters, we design a preprocessing pipeline. Angle and distance normalization, frame division, and point re-sampling are the steps of the pipeline in the mentioned order.

\subsubsection{Angle and distance normalization}
To reduce the effect of angle and distance of the participant w.r.t. the antenna center-line on the accuracy, data normalization is performed. To do so, we use affine-geometric transformation matrices to rotate and translate the data to the reference point (1.5m distance and 0 angle in Pantomime and 1m distance and 0 angle in RadHAR).

\subsubsection{Frame divider}
To study the effect of number of frames on system accuracy, complying with the temporal order of points, we distribute them in different number of frames (2, 4, 8, 16, 32, 64). Assume $S$ is the desired number of frames and $n$ is the total number of points in the recorded gesture. We consider first $n/S$ points as the first frame, second $n/S$ points as the second frame and so on.

\subsubsection{Point re-sampling}
To study the effect of number of points in each frame on the system performance, we employ a density-based re-sampling strategy introduced by Cohen et al. \cite{cohen2006learning} to preserve the spatial structure while fixing the number of points in each frame.
Considering $n/S$ as the desired number of points in each frame, to reduce the number of points we use $\mathcal{K}$-means algorithm and set $\mathcal{K}$ equal to $n/S$ and select the centroids of the clusters as the points in the frame. To increase the number of points in the frame to $n/S$, we iteratively apply \gls{ahc} and add the centroids of the clusters as new points to the frame until we have the desired number of points.

\subsection{Data Augmentation}
Different data augmentation techniques are applied to improve the generalizability of the system in terms of different angles, distances, and scales. We apply the following augmentations to each batch during the training phase:
\begin{itemize}
    \item Random translation up to 10cm
    \item Random scaling between 0.8 to 1.25
    \item Random point-wise translation (jitter) based on a Gaussian distribution with $\mu=0$ and $\sigma=0.01$
    \item Random clipping of 0.03m
    \item Random shuffling of the point cloud representation preserving the spatial and temporal features
\end{itemize}

\subsection{Training and Inference}
The infrastructure used for training and inference phases has 64GB of RAM and is equipped with a Tesla V100 16GB GPU. The model is implemented using PyTorch \cite{paszke2019pytorch} and PyTorch Geometric \cite{fey2019fast}.
We utilize early stopping mechanism in the training phase with a patience of 100 epochs. To do so, if no improvement on validation set accuracy is observed within the patience period, training is stopped and the best model is saved.
The loss function used for training the model is cross-entropy between class scores and one-hot encoded labels. To minimize this loss function, we use Adam Optimizer \cite{kingma2014adam} with a step-decay strategy to decrease learning rate:
\begin{equation}
    L_{r} = L_{init} \cdot d_{r} ^ {\lfloor \frac{e} { e_{r}}\rfloor} 
    \label{eq:step_decay}
\end{equation}
where $L_{r}$ is the learning rate used at each epoch, $L_{init}$ is the initial value of the learning rate, $d_{r}$ is the drop rate after every $e_{r}$ epochs, $e$ is the current epoch and $\lfloor\cdot\rfloor$ is the floor operator. In our setup $L_{init}$ is $0.001$, $d_{r}$ is $0.5$, and $e_{r}$ is $20$.

\subsection{Real-time Implementation}
We implement \system~for real-time gesture recognition on Raspberry PI 4 device with 8GB RAM, as an example embedded device with constrained computing resources.

\begin{algorithm}
\SetAlgoLined
\KwResult{Recognized gesture}
 $frame\_list=[]$, $min\_frames=2$, $idle\_frame\_count=0$,
 $idle\_frame\_delimiter=10$,
 $idle\_frame\_threshold=3$\;
 \While{Receive $frame\_data$ from Radar}{
  \If{len($frame\_list$) $\geq$ $min\_frames$ and $idle\_frame\_count$ $\geq$ $idle\_frame\_delimiter$} {
    preprocess($frame\_list$)\;
    perform\_recognition($frame\_list$)\;
   }
   \If{len($frame\_data$) $\leq$ $idle\_frame\_threshold$}{
    $idle\_frame\_count += 1$\;
    $continue$\;
   }
   $idle\_frame\_count = 0$\;
   append $frame\_data$ to $frame\_list$\;
 }
 \caption{Real-time recognition algorithm}
 \label{algo:real_time_recognition}
\end{algorithm}
For recognizing gestures in real-time, we develop an algorithm which uses \model~model as classifier. We categorize each captured frame into two sets of \textit{active frame} and \textit{idle frame}. \textit{Idle frames} are frames in which no notable movement is observed and the rest are considered as \textit{active frame}. In Algorithm \ref{algo:real_time_recognition}, we use a set of consecutive idle frames as a delimiter for different gestures. A similar approach is employed in different gesture recognition systems like DoubleFlip \cite{ruiz2011doubleflip} and WristRotate \cite{kerber2015wristrotate} or even in voice assistants, e.g., \cite{masina2020investigating}. Gesture recognition is performed whenever a minimum number of active frames are identified. Thresholds for minimum active frames, gesture delimiter, and maximum number of points for idle frames are denoted as $min\_frames$, $idle\_frame\_delimiter$, $idle\_frame\_threshold$, respectively and tuned empirically.

The real-time recognition algorithm is implemented on a Raspberry PI 4 with a connected IWR1443 Radar responsible for sensing the human movement (see section \ref{sec:point_cloud_generation}). A Cortex-R4F built-in micro-controller is employed in the radar and the universal asynchronous receiver-transmitter protocol realizes data transfer. We configure the device to capture frames at a rate of 30 fps with a range resolution of 0.047 m, a velocity resolution of 0.87 m/s, and a maximum velocity of 6.9 m/s up to a maximum range of 5 m. The starting frequency is 77GHz and our selected range resolution dictates a bandwidth of 3.19GHz.

\section{Evaluation}
This section presents the performance evaluation of the \model~model and \system~system in terms of recognition accuracy and time complexity.
\subsection{Datasets}
%
\begin{table}
\centering
\begin{tabular}{@{}lcc@{}}
\toprule
                             & \textbf{Pantomime}   & \textbf{RadHAR} \\ \midrule
\textbf{Participants}        & 41                   & 2                \\
\textbf{Number of classes}     & 21                   & 5                \\
\textbf{Max. range (m)}      & 5 m                  & ~ 1.5 m            \\
\textbf{Environments}        & 5                    & 1                 \\
\textbf{Frame rate (fps)}    & 30                   & 60                \\
\textbf{Training samples}    & 7000                 & 12097             \\ \bottomrule
    \addlinespace[0.5em]
\end{tabular}
\caption{Comparison of the two datasets.}
\label{tab:two_datasets}
\end{table}

For evaluation purpose, we use two radar generated point cloud datasets: Pantomime \cite{palip2021pantomime} and RadHAR \cite{singh2019radhar}.
The comparison between two datasets is shown in Table~\ref{tab:two_datasets}. Both  datasets were acquired using a 77 GHz IWR1443 millimeter wave radar. The gestures in Pantomime are divided into three sets: \textit{Easy} (9 classes), \textit{Complex} (12 classes) and \textit{All} (21 classes) based on the execution difficulty. The \textit{Easy} set comprises single-hand gestures that are easy to perform and remember. The \textit{Complex} set comprises bimanual, linear, and circular gestures. Finally, \textit{All} consists of gestures from both sets. The training data in RadHAR is collected from one anchor position of 1.5m, whereas the training data in Pantomime is collected from 4 anchor positions between 1.5 to 5m.
For evaluating the model on Pantomime and RadHAR datasets, we employ the same train, validation, test splits provided by Pantomime and RadHAR authors, respectively.

\subsection{Hyperparameter Tuning}
\label{sec:hypertuning}
\begin{figure*}
    \centering
    \begin{minipage}{.73\linewidth}
    \subfigure[]{
        \includegraphics[width=.32\textwidth]{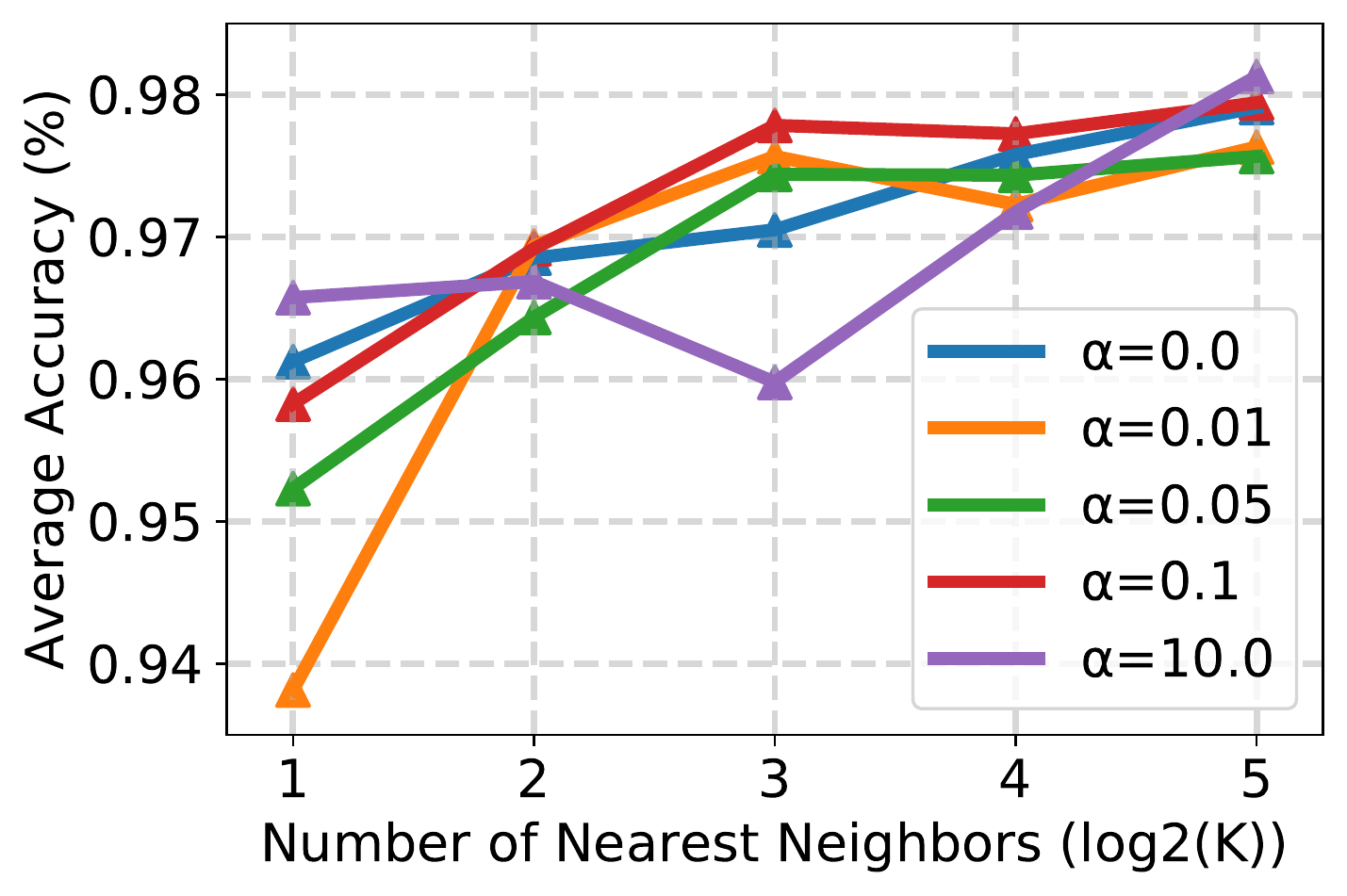}%
        }
    \subfigure[]{
        \includegraphics[width=.32\textwidth]{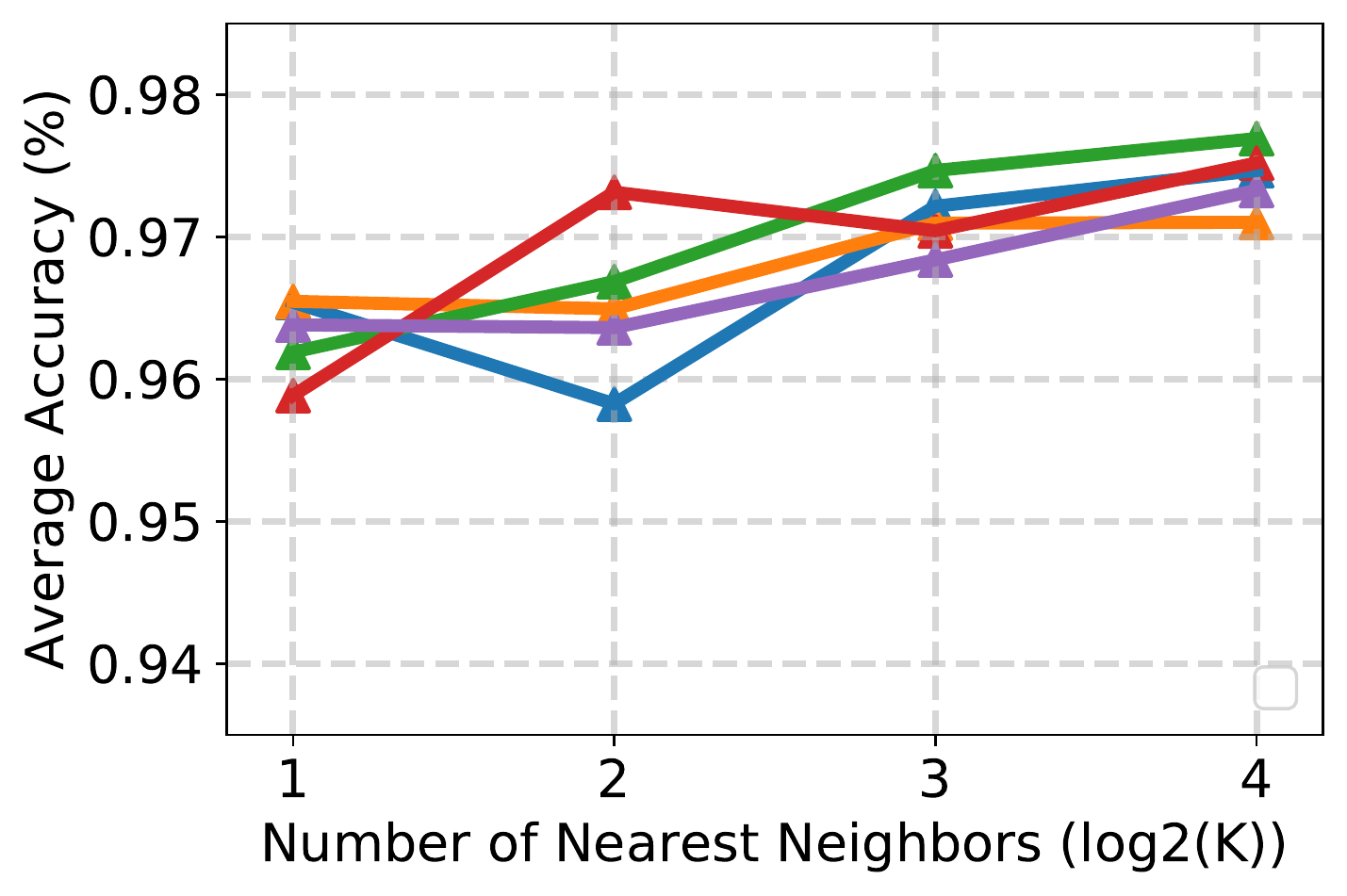}%
        }
    \subfigure[]{
        \includegraphics[width=.32
        \textwidth]{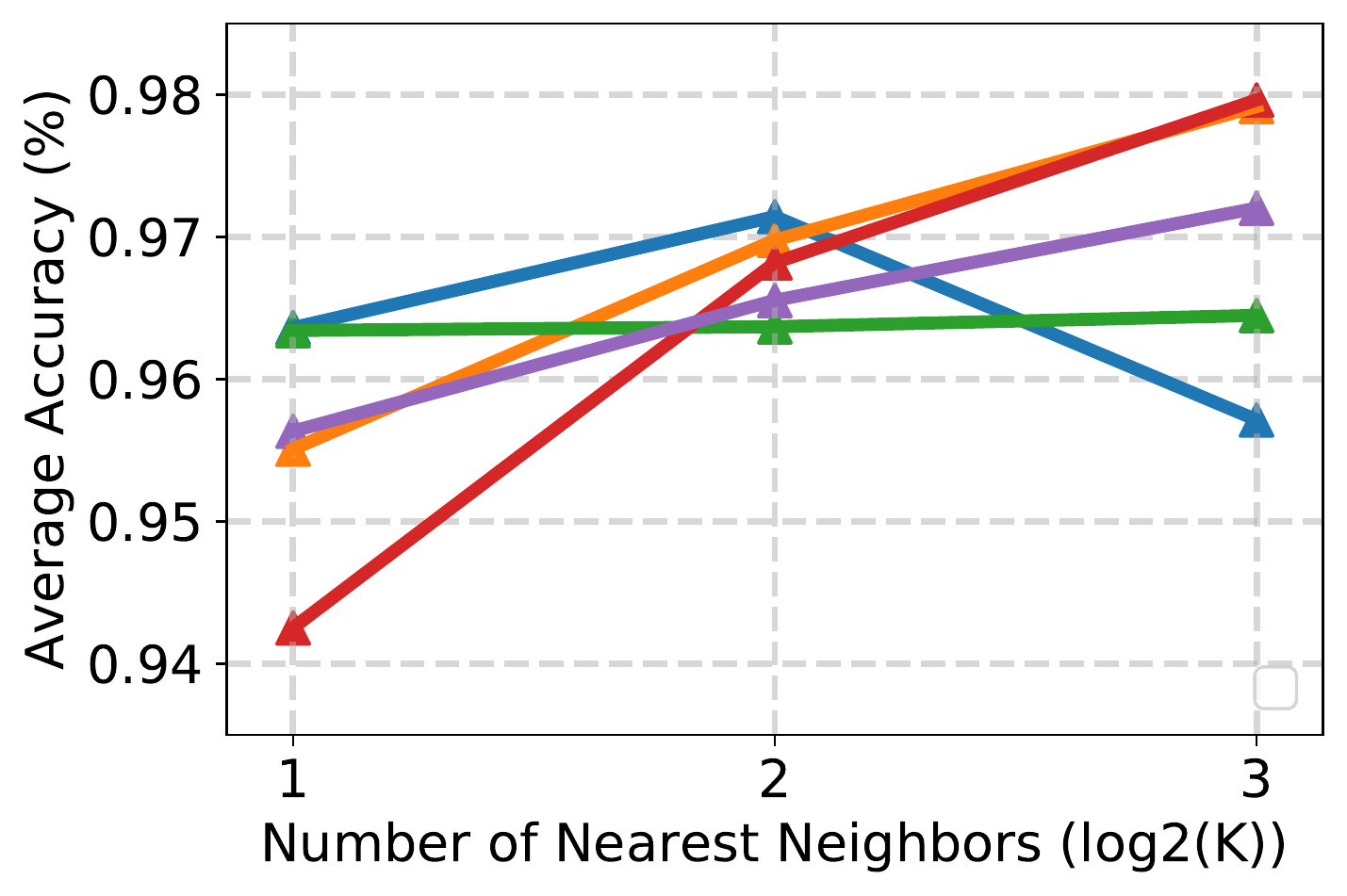}%
        }
    \caption{The effect of hyper parameters. 
    (a) One, (b) two  and (c) three layers.}
    \label{fig:hyperparameters}
    \end{minipage}
    \hfill
    \begin{minipage}{.24\linewidth}
    \includegraphics[width=\linewidth]{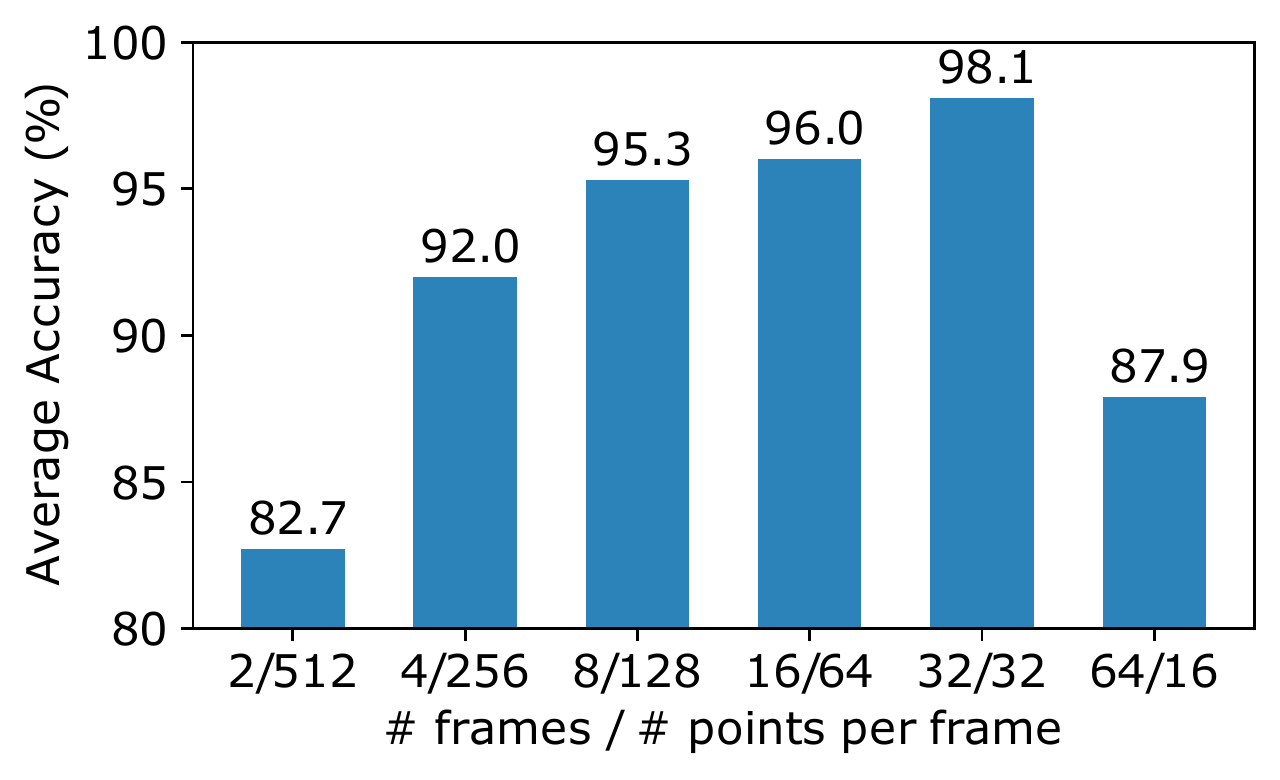}
    \caption{The impact of the \# of frames and points per frame on avg. accuracy.}
    \label{fig:frame_point_effect}
    \end{minipage}
\end{figure*}

The results of hyperparameter tuning of the model on Pantomime validation dataset are illustrated in Fig. \ref{fig:hyperparameters}. In order to tune each point's neighbors number ($k$) and the value of $\alpha$ in \proposedknn~and the number of \layername~layers, different combinations of parameters are used to train the model and test it on the validation set. According to the cross-validation process, best results are obtained using one layer of \layername~with $k=32$ and $\alpha=10$. Increasing the complexity of the model by adding more layers does not contribute to the accuracy of the model. Although, increasing $k$ leads to a more complex model since \proposedknn~generates denser graphs (see Fig. \ref{fig:sys_archi}), in most of the cases the accuracy is enhanced as demonstrated in Fig. \ref{fig:hyperparameters}. In general, no clear trend is observable when it comes to changing $\alpha$ indicating that performance of different $\alpha$ values is not independent from values of $k$ and the number of layers.

We choose two sets of hyperparameters: \model~model with $k=32$ and $\alpha=10$, the best performing one in terms of accuracy, and \lighmodel~(Tesla-Vanilla) model with $k=2$ and $\alpha=10$, reasonably accurate but faster than \model~in terms of prediction time.

In Fig. \ref{fig:frame_point_effect} the impact of the number of frames and the number of points in a frame on the average accuracy is evaluated. Six settings of different combinations of the number of frames and the number of points per frame are considered while keeping the total number of points (=number of frames $\times$ \text{number of points per frame}) in each gesture is a constant (1024).  Increasing the number of frames up to 32, improves the accuracy. However, adding more frames than 32 to the gesture decreases the accuracy indicating both number of frames and number of points in each frame play important roles in performance of the system.

Additionally, to illustrate the effect of self-attention mechanism on the performance of the model, we train the \model~without the self-attention mechanism on the training set of the Pantomime dataset. The overall accuracy of the trained \model~ without self-attention on the validation set is $\textbf{95.2\%}$ ($\textbf{3\%}$ drop compared to the model with self-attention) indicating the positive effect of self-attention in improving the performance of the model.

\subsection{Classification Results}
\subsubsection{Overall Results on Pantomime dataset}
\label{sec:classification_results}
In Table \ref{table:soa_comp_pantomime}, the performance of \model~and \lighmodel~on Pantomime dataset is compared to baseline models of \textit{PointNet} \cite{qi2017pointnet}, \textit{PointNet++} \cite{qi2017pointnet2}, \textit{O\&H} \cite{owoyemi2018spatiotemporal}, \textit{PointGest} \cite{salami2020motion}, \textit{RadHar} \cite{singh2019radhar}, \textit{PointLSTM} \cite{min2020pointlstm}, \textit{Pantomime} \cite{palip2021pantomime}, and \textit{\gls{dec}} \cite{wang2019dynamic}. In PointNet, PointNet++, and \gls{dec}, the frames are aggregated through time dimension into a single frame representing the whole gesture, since they are designed to classify static point clouds. While the rest of the models aim to classify motion point clouds. Moreover, from input data representation perspective, O\&H and RadHar work on voxels whereas the rest of them directly operate on point clouds. In case of Pantomime and PointGest, gestures are represented with 8 frames (same number as in the original papers) since they are computationally demanding and not feasible to run with more frames on the same infrastructure. As illustrated in Table \ref{table:soa_comp_pantomime}, our \model~model outperforms all baselines in every category, in terms of accuracy and \gls{auc} (measuring the discriminative capability of models). Additionally, \model~model increases the accuracy of state of the art by $\textbf{0.9}\%$, $\textbf{4.2}\%$, $\textbf{3.1}\%$ in Easy, Complex, and All settings, respectively, as well as achieving $\textbf{100}\%$ \gls{auc} in both Complex and All. Furthermore, \lighmodel~model performs rather efficiently compared to baselines and \model, ranking 2nd on Complex and All and 3rd (only $0.1\%$ behind 2nd) on Easy when it comes to accuracy.

\begin{table}[t]
	\centering
	\begin{tabular}{l *6c}
		\toprule
		  & \multicolumn{2}{c}{\textbf{\easy}}    
		  & \multicolumn{2}{c}{\textbf{\complex}} 
		  & \multicolumn{2}{c}{\textbf{\all}}     
		\\
		\cmidrule(l){2-3}
		\cmidrule(l){4-5}
		\cmidrule(l){6-7}
		\textbf{Model}
		  & \textbf{Acc.} 
		  & \textbf{\gls{auc}}
		  & \textbf{Acc.}
		  & \textbf{\gls{auc}}
		  & \textbf{Acc.}
		  & \textbf{\gls{auc}}
		\\ \midrule
		PointNet 
		  & 79.7
		  & 98.4
		  & 82.5
		  & 98.7
		  & 81.6
		  & 99.4
		\\
		PointNet++ 
		  & 79.7
		  & 98.1
		  & 84.9
		  & 99.0
		  & 83.6
		  & 99.4
		\\
		O\&H 
		  & 77.7
		  & 96.0
		  & 83.2
		  & 98.1
		  & 79.1
		  & 97.9
		\\
		PointGest 
		  & 83.3
		  & 98.5
		  & 88.4
		  & 99.5
		  & 86.3
		  & 99.4
		\\
		RadHAR 
		  & 91.6
		  & 98.9
		  & 94.3
		  & 99.6
		  & 89.9
		  & 99.5
		\\
		PointLSTM
		  & 85.1
		  & 99.1
		  & 92.1
		  & 99.8
		  & 90.7
		  & 99.7
		\\
		Pantomime
		  & 96.6
		  & \textbf{99.8}
		  & 95.1
		  & 99.8
		  & 95.0
		  & 99.9
		\\
		\gls{dec} 
		  & 81.9
		  & 98.2
		  & 89.1
		  & 99.3
		  & 86.0
		  & 99.4
		\\
		\bottomrule
		\addlinespace[0.5em]
		\lighmodel{}
		  & 96.2
		  & 99.7
		  & 99.1
		  & 100
		  & 96.6
		  & 99.9
	    \\
		\model{}
		  & \bf 97.5
		  & \bf 99.8
		  & \bf 99.3
		  & \bf 100
		  & \bf 98.1
		  & \bf 100
		\\ \bottomrule
		\addlinespace[0.5em]
	\end{tabular}
	\caption{Comparison with the state of the art on the Pantomime dataset. Both Acc. (Accuracy) and \gls{auc} are reported in percentages.
	The best results per column are denoted in bold typeface.}
	\label{table:soa_comp_pantomime}
\end{table}

\subsubsection{Different Environments}
We also evaluated \model~model with different environments on Pantomime dataset, comparing to the closest competitor. Following the same approach as \cite{palip2021pantomime}, the model is trained on data acquired in Open and Office settings and tested against five different environments reported in Table \ref{table:pantomime_different_settings}. We manage to improve accuracy up-to $10\%$ in all environments except for Open. This arises from the fact that the frames in cluttered environments like Through-wall are sparser compared to less cluttered environments , e.g., Open. Therefore, the spatial distribution of the frames in the train set is different from that of the test set. Consequently, the models capturing spatial features and fusing them through \gls{lstm} layers i.e. Pantomime, fail to generalize well (see section \ref{sec:point_cloud_properties}). On the contrary, \model~model, recognizes gestures based on their temporal structures which leads to a more robust prediction in unseen environments.

\subsubsection{Different Speeds} In addition, the effect of gesture speed is illustrated in Table \ref{table:pantomime_different_settings}. The models are trained on gestures performed with Normal speed and tested on Slow, Normal, and Fast speeds. \model~model outperforms Pantomime in Normal and Fast articulation speeds. However, in setting \textit{Slow}, we are behind state of the art.

\begin{table}[t]
	\centering
	\begin{tabular}{l *6c}
		\toprule
		  & \multicolumn{2}{c}{\textbf{Pantomime}}    
		  & \multicolumn{2}{c}{\textbf{Proposed}} 
		\\
		\cmidrule(l){2-3}
		\cmidrule(l){4-5}
		\textbf{Setting}
		  & \textbf{Acc.} 
		  & \textbf{\gls{auc}}
		  & \textbf{Acc.}
		  & \textbf{\gls{auc}}
		\\ \midrule
		Factory 
		  & 89.11
		  & 99.79
		  & \bf 97.14
		  & \bf 99.96
		\\
		Restaurant 
		  & 81.13
		  & \bf 98.84
		  & \bf 82.14
		  & 98.19
		\\
		Office 
		  & 93.40
		  & 99.86
		  & \bf 97.14
		  & \bf 99.94
		\\
		Open 
		  & \bf 96.12
		  & \bf 99.94
		  & 94.36
		  & 99.88
		\\
		Through-wall 
		  & 64.43
		  & 97.24
		  & \bf 74.64
		  & \bf 98.51
		\\
		\bottomrule
		\addlinespace[0.3em]
		Slow
		  & \bf 85.00
		  & \bf 99.33
		  & 76.19
		  & 98.69
		\\
		Normal
		  & 94.05
		  & 99.90
		  & \bf 95.95
		  & \bf 99.95
		\\
		Fast 
		  & 92.14
		  & 99.68
		  & \bf 94.28
		  & \bf 99.87
		\\ \bottomrule
		\addlinespace[0.5em]
	\end{tabular}
	\caption{Comparison with Pantomime model (the closest competitor) on different settings of the Pantomime dataset. The best Acc. and \gls{auc} per row are denoted in bold typeface.}
	\label{table:pantomime_different_settings}
\end{table}
\begin{table}[t]
	\centering
	\begin{tabular}{l *6c}
		\toprule
		\textbf{Model}
		  & \textbf{Acc.}
		  & \textbf{\gls{auc}}
		\\ \midrule
		\gls{svm}
		  & 63.74
		  & -
		\\
		\gls{mlp}
		  & 80.34
		  & -
		\\
		Bi-directional \gls{lstm}
		  & 88.42
		  & -
		\\
		RadHAR
		  & 90.47
		  & -
		\\
		PointLSTM
		  & 94.11
		  & 98.70
		\\
		Pantomime
		  & 94.19
		  & 99.65
		\\
		\gls{dec}
		  & 96.24
		  & 99.62
		\\
		\bottomrule
		\addlinespace[0.5em]
		\lighmodel{} (ours)
		  & 95.49
		  & 99.48
	    \\
		\model{} (ours)
		  & \bf 96.97
		  & \bf 99.75
		\\ \bottomrule
		\addlinespace[0.5em]
	\end{tabular}
	\caption{Comparison with the state of the art on the RadHar. The Accuracy is reported in percentages. The performance of \textit{\gls{svm}}, \textit{\gls{mlp}}, \textit{Bi-directional \gls{lstm}}, and \textit{RadHar} are reported from \cite{singh2019radhar}.
	The best results per column are denoted in bold typeface.}
	\label{table:soa_comp_radhar}
\end{table}
\begin{figure}
    \centering
    \subfigure[]{
        \includegraphics[width=.48\columnwidth]{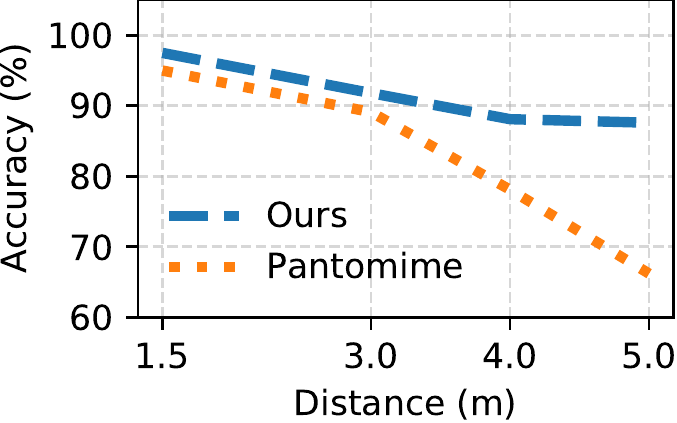}%
        }
    \subfigure[]{
        \includegraphics[width=.48
        \columnwidth]{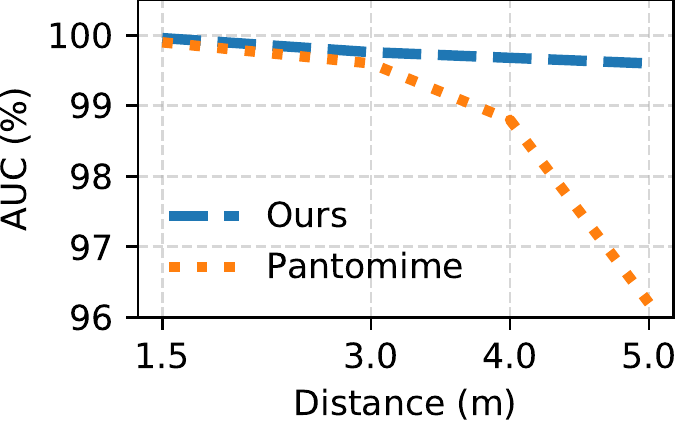}%
        }
    
        \subfigure[]{
        \includegraphics[width=.48\columnwidth]{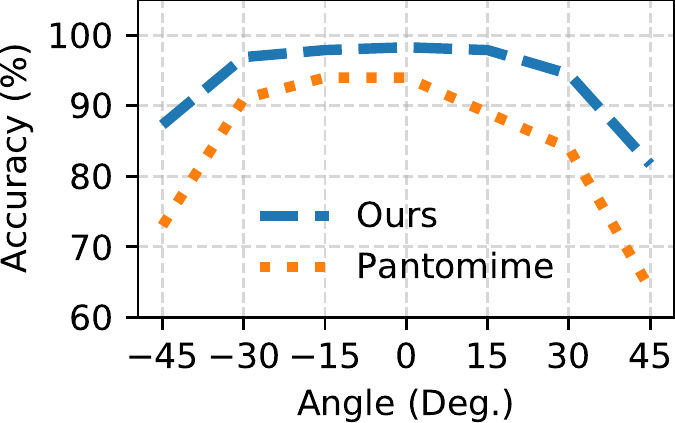}%
        }
    \subfigure[]{
        \includegraphics[width=.48
        \columnwidth]{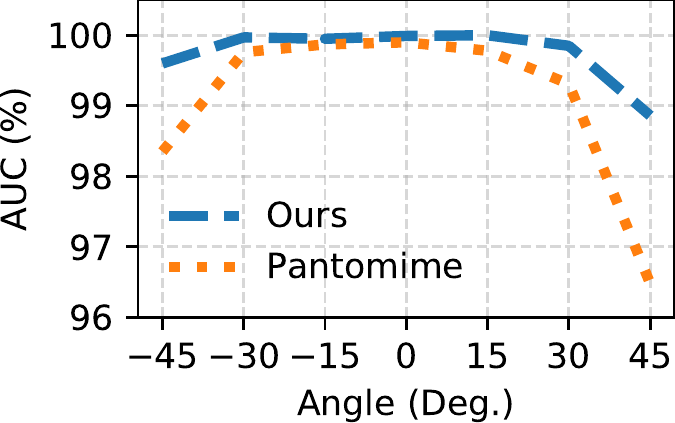}%
        }
     \caption{Comparison between Pantomime and \model~models. (a), (b) on different   distances  and (c), (d) on angles  in terms of accuracy and \gls{auc}}

    \label{fig:robustness}
\end{figure}

\subsubsection{Different Distances and Angles} For measuring the robustness of the prediction against the position of the participant w.r.t. radar, we compared the performance of \model, on different angles and distances. As shown in Fig. \ref{fig:robustness}, our \model~model outperforms Pantomime in every setting of angle and distance in terms of both accuracy and \gls{auc}. When it comes to extreme setups i.e. 5m distance, $-45^\circ$ and $45^\circ$ angels, \model~is significantly ahead of Pantomime improving the accuracy up to $21\%$. Furthermore, with the increase of distance, the performance drop in our \model~is less than $10\%$, whereas Pantomime degrades in accuracy with an exponential rate (almost $30\%$). Given the change in the distribution of point clouds in different configurations of the radar (see section \ref{sec:point_cloud_properties}), Pantomime fails to generalize since it extracts spatial features from each frame, fusing them to identify temporal pattern. However, \model~recognizes gestures based on the temporal graph which is more robust to angle and distance.

\subsubsection{Overall results on RadHAR dataset} In Table \ref{table:soa_comp_radhar}, the results of different models on RadHar dataset are illustrated. \gls{svm}, \gls{mlp}, \textit{Bi-directional \gls{lstm}}, and \textit{RadHar} use voxels as input. As shown in Table \ref{table:soa_comp_radhar}, \model~model outperforms baselines in both measures of accuracy and \gls{auc}. Moreover, \lighmodel~ranks third in the table in terms of both accuracy and \gls{auc}.

\begin{figure*}[t]
    \begin{minipage}[t]{.45\linewidth}
        \centering
        \includegraphics[width=0.9\linewidth]{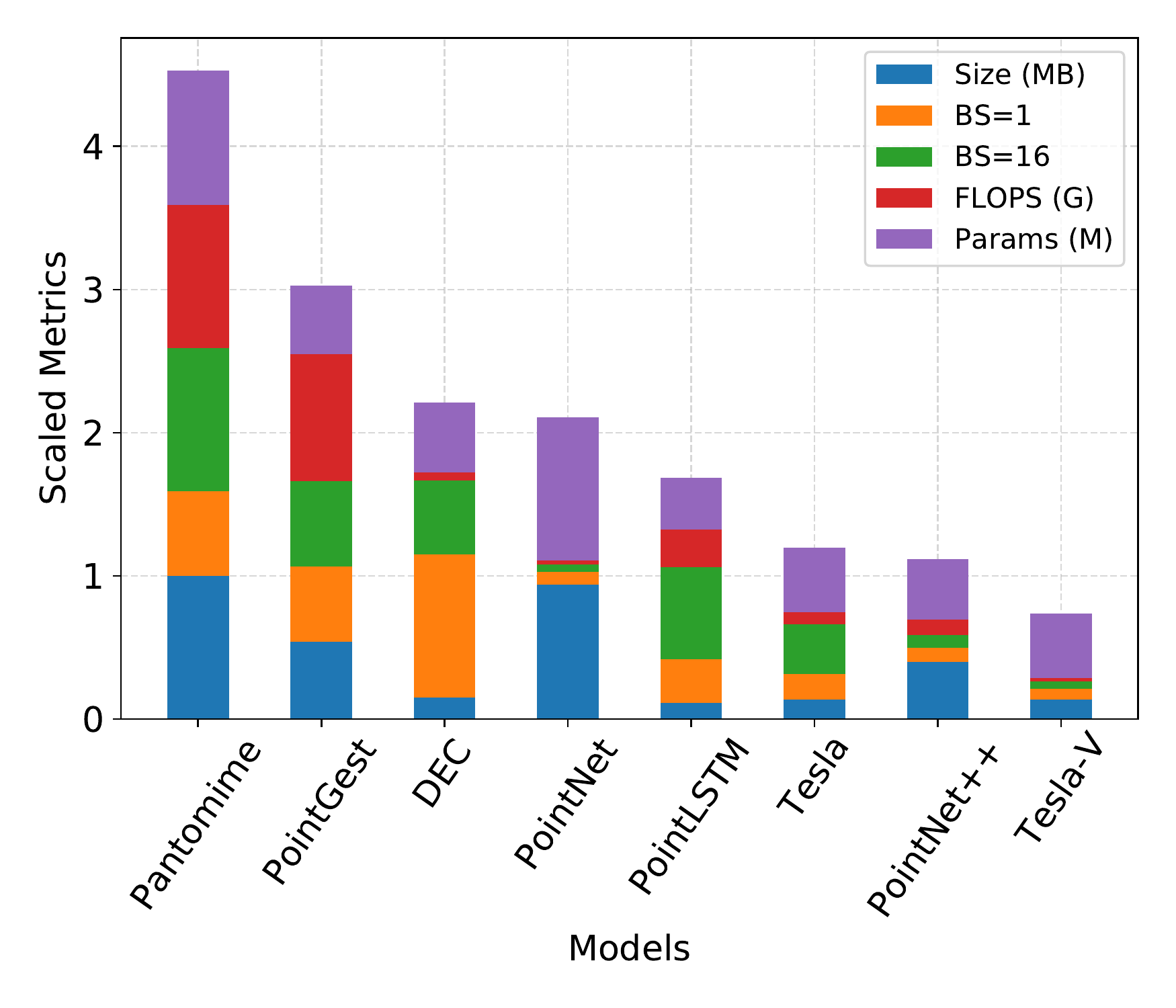}
        \caption{Scaled model complexity comparison. Each metric is scaled between 0 and 1. Size: model size, BS: average inference time a batch size, Params: trainable parameters}
        \label{fig:computational_complexity}
    \end{minipage}
    \hfill
    \begin{minipage}[t]{.45
    \linewidth}
        \centering
        \includegraphics[width=0.9\linewidth]{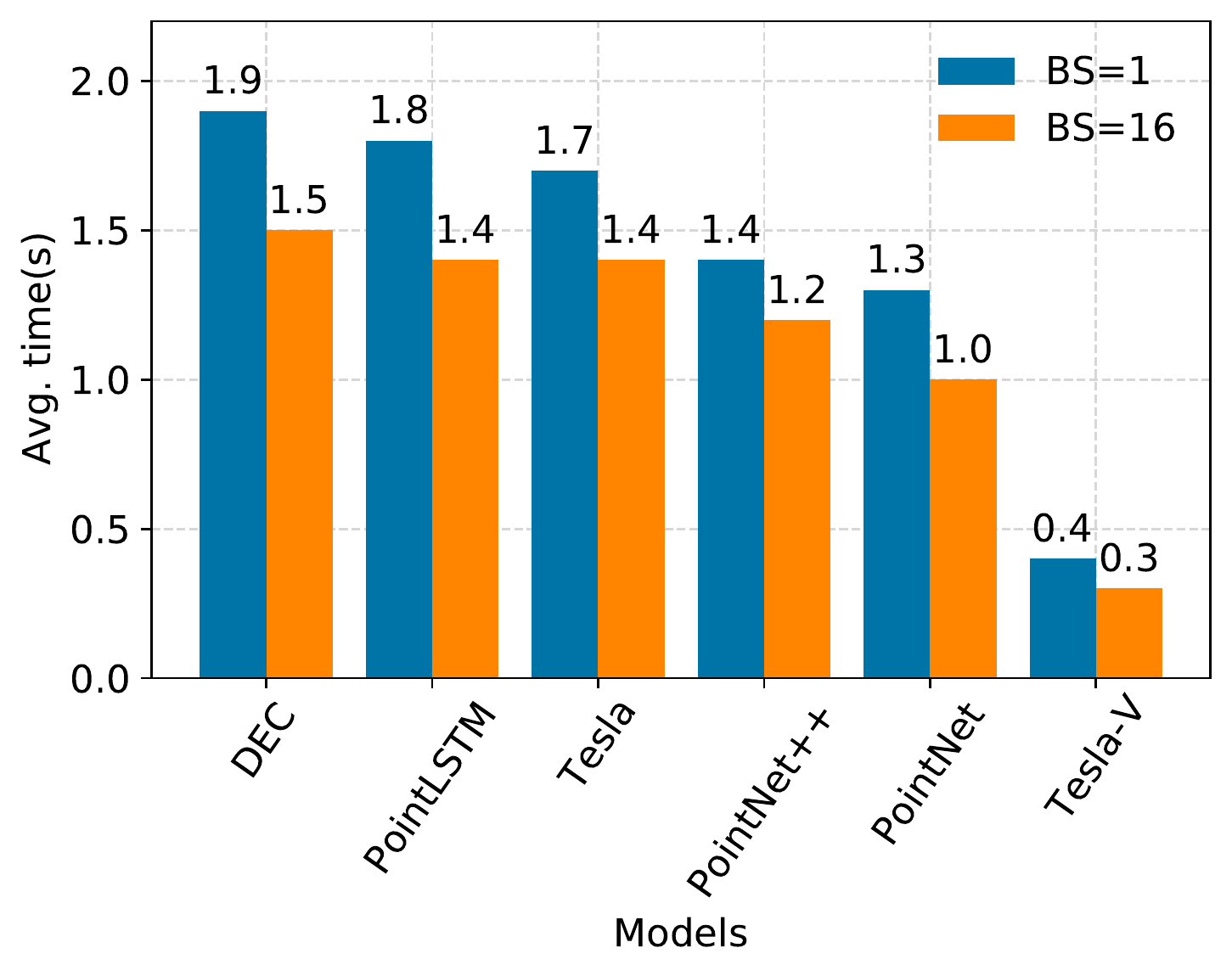}
        \caption{Average inference time per gesture of proposed models on a Raspberry PI device}
        \label{fig:raspberry_pi_inference}
    \end{minipage}
\end{figure*}
\subsection{Time Complexity Results}
\label{sec:time_complexity_results}

In Fig. \ref{fig:computational_complexity}, time complexity comparison between \model~and \lighmodel~and baselines on Pantomime dataset on a Tesla V100 \gls{gpu} with 16GB of memory is presented. To evaluate the efficiency of the model, we measure four different metrics of \textit{average inference time},  \textit{\glspl{gflop}}, \textit{number of trainable parameters}, and \textit{size} of the trained model. For measuring inference time, two settings of batch size 1 and 16 were considered and the average time of 10 forward passes were gathered after warming up the infrastructure by running a few batches. Each category of measurements in Fig. \ref{fig:computational_complexity} are scaled based on the maximum value of the category. According to Fig. \ref{fig:computational_complexity}, \lighmodel~model has the lowest aggregate complexity among all the models. Furthermore, \model~model, which is the best performing one in terms of accuracy, ranks 3rd in total, just behind PointNet, a model that does not take into account the temporal dependency, therefore, having a much lower accuracy. Compared to the most accurate competitor (Pantomime), \lighmodel~is 18 and 8 times faster in inference with batch sizes 16 and 1 respectively; and 40 times computationally efficient in terms of \glspl{gflop}. In addition, computationally closest competitor is PointNet which has almost the same inference time and \glspl{gflop} while falling behind \lighmodel~by 16.5\% when it comes to recognition accuracy.

\subsection{Real-time Implementation Evaluation}

\begin{figure}
    \centering
\includegraphics[width=0.7\columnwidth]{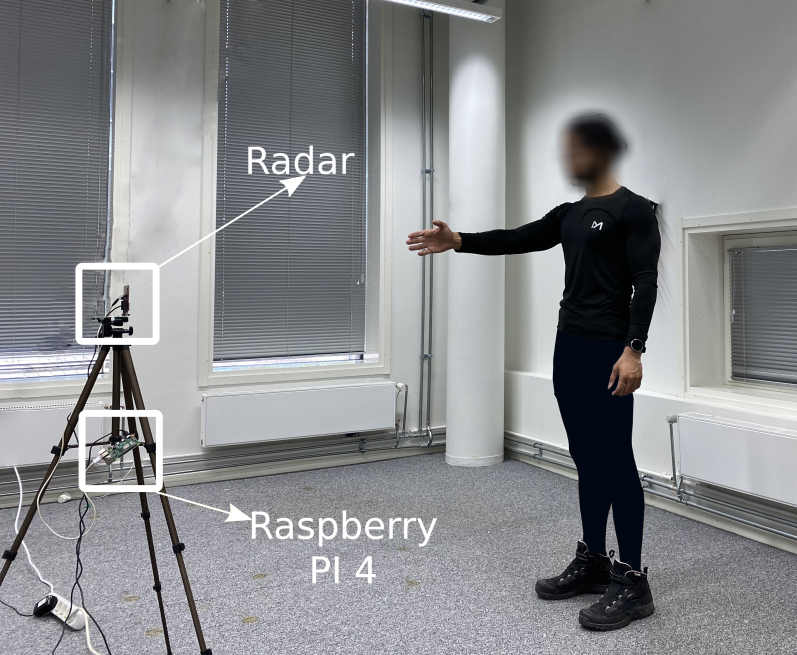}
    \caption{\system~setup for real-time evaluation using an IWR1443 radar and Raspberry PI 4 device}
    \label{fig:real_time_setup}
\end{figure}
In this part we evaluate inference time and performance of the model on a Raspberry PI 4 device with 8GB of RAM. The setup for the real-time testbed is shown in Fig. \ref{fig:real_time_setup}.

\subsubsection{Inference time}
In Fig. \ref{fig:raspberry_pi_inference}, the average inference time on a Raspberry PI 4 with 8GB of memory with batch sizes of 1 and 16, for \gls{dec}, PointLSTM, \model, PointNet++, PointNet, and \lighmodel~are illustrated. Since, Pantomime and PointGest require more than 8GB of RAM, their implementation on Raspberry PI is not feasible. Among the implemented models, \lighmodel~is able to predict gestures in 0.4s and 0.3s with batch sizes 1 and 16 respectively, making it the only model predicting a gesture in less than half a second. As a result, \lighmodel is the only suitable model for integration into the real-time gesture recognition interface \system, since inference time of more than one second in the case of all other models, is not fast enough for real-time user experience.

\subsubsection{Performance}
To evaluate the performance of the purposed gesture recognition system, \system, the pipeline shown in Fig. \ref{fig:tesla_rapture} is implemented. The prediction model in this system is \lighmodel~(the only model with inference time of less than one second) and the inference is done on a Raspberry PI device connected to an IWR1443 radar for gathering gestures.

Since Pantomime dataset does not have a class for rejecting gestures (\textit{no-gesture} class), 2 hours of moving point clouds in which there were random movements of participants as well as idle frames were recorded. The training data from Pantomime dataset was combined with \textit{no-gesture} samples to train the model. As a result, to train \lighmodel~for the real-time system, we used 22 classes including a \textit{no-gesture} class to reject the gestures that do not belong to the gesture set of Pantomime dataset.

In the evaluation phase of the \system, we asked 5 participants to perform each class of gesture for 10 times as well as doing random movements in front of the radar (like walking, staying idle, and doing some random gestures other than the original gesture set). To do so, we showed gesture videos to participants and asked them to perform each gesture a few times before the actual evaluation. In the evaluation round, we showed the name and the schematic view of the gesture in a random order on the screen and the participant performed the gesture. As shown in algorithm \ref{algo:real_time_recognition}, we use a few idle frames as a gesture delimiter. Consequently, between each gesture there was one second gap. The overall accuracy of the real-time system is \textbf{90.53\%} and the false-alarm rate is \textbf{4.4\%}.

\section{Discussion}
\fakeparagraph{\textbf{Gesture Recognition}} Introducing \system~system, as a fast and accurate gesture recognition interface is a step forward in human-computer interaction scenarios for integration with many off-the-shelf devices. Given the robustness of the system in different environments, angels, and distances as well as real-time performance, \system~system can be incorporated into a wide range of applications e.g., smart-homes, vehicular settings, and human-robot interaction. Furthermore, the model can be trained on a customized set of gestures and deployed on \system~for a specific real-time application.

\fakeparagraph{\textbf{Speed vs. Accuracy}} High performance of \model~makes it suitable for sensitive applications in which the accuracy cannot be compromised. However, this performance comes with a cost of slower recognition speed. To address this issue, we introduced \lighmodel, a faster prediction model, with only $1.5\%$ drop in accuracy while performing inference 3 times faster than \model. Thus, \model~and \lighmodel~cover a wide range of applications with different speed-accuracy requirements.

\fakeparagraph{\textbf{Egocentric Applications}} Due to the computational efficiency and robustness to different environments and angles, \system~system can be extended to scenarios where egocentric gestures should be recognized on constrained devices. \model~prediction model can be modified to adapt to new applications for wearable devices, e.g., Microsoft HoloLens\footnote{https://www.microsoft.com/en-us/hololens}. Currently, HoloLens 2 captures hand gestures using RGB-D sensors. Given the benefits of radars over RGB-D cameras (see section  \ref{sec:related_work}), integration of \system~with HoloLens improves the performance of hand gesture recognition which is one of the main interaction mechanisms of this device.

\fakeparagraph{\textbf{\model~on Dense Point Clouds}} We trained and evaluated \model~on SHREC-28 dataset\cite{de20173d}, a set of dense gestures collected using a depth camera. The proposed model achieves $81.5\%$ accuracy while the state of the art (PointLSTM) has $94.7\%$ accuracy suggesting that \model~fails to capture spatial structures in each frame effectively which is vital for dense point cloud processing. Since \model~aims at recognizing gestures from mmWave radar generated point clouds, highly sparse compared to that of other devices (see section \ref{sec:comparison_with_rgb_d_point_cloud}), capturing spatial features of each frame does not contribute to the performance. Our approach outperforms PointLSTM (state of the art model on SHREC-28), with a margin of up to $12.4\%$ accuracy on mmWave radar generated point clouds (see section \ref{sec:classification_results}).

\fakeparagraph{\textbf{Future Work}} While in this work, we introduced \proposedknn~to reflect the temporal dependency in graph generation, the graph is still being created statically using \gls{knn} algorithm. \gls{rl}, imitating the cognitive reward based learning process, enhances the graph according to the accuracy of the classification model. Therefore, dynamic graph generation using \gls{rl} is one possible direction for improving the temporal graph.

\section{Conclusion}
In this work, we proposed \system, a real-time gesture recognition interface based on mmWave radar generated sparse point clouds. In doing so, we designed \proposedknn~to implicitly reflect the temporal evolution of gestures in a temporal graph on which the proposed attention-based \gls{mpnn} is applied to recognize gestures. Moreover, we presented two versions of \model~and \lighmodel~employing the mentioned strategy. Our results show that \system~enhances the accuracy up to 21\% in extreme settings while reducing the prediction time by a magnitude of 8 and computational complexity (\glspl{gflop}) by almost 40 times compared to the most accurate competitor.
\ifCLASSOPTIONcaptionsoff
  \newpage
\fi

\section*{Acknowledgment}
We thank the anonymous referees for the constructive feedback provided. Part of the calculations presented above were performed using computer resources within the Aalto University School of Science “Science-IT” project.

This project has received funding from the European Union’s Horizon 2020 research and innovation programme under the Marie Skłodowska-Curie Grant agreement No. 813999.

\bibliographystyle{IEEEtran}
\bibliography{references}

\begin{thebibliography}{10}
\providecommand{\url}[1]{#1}
\csname url@samestyle\endcsname
\providecommand{\newblock}{\relax}
\providecommand{\bibinfo}[2]{#2}
\providecommand{\BIBentrySTDinterwordspacing}{\spaceskip=0pt\relax}
\providecommand{\BIBentryALTinterwordstretchfactor}{4}
\providecommand{\BIBentryALTinterwordspacing}{\spaceskip=\fontdimen2\font plus
\BIBentryALTinterwordstretchfactor\fontdimen3\font minus
  \fontdimen4\font\relax}
\providecommand{\BIBforeignlanguage}[2]{{%
\expandafter\ifx\csname l@#1\endcsname\relax
\typeout{** WARNING: IEEEtran.bst: No hyphenation pattern has been}%
\typeout{** loaded for the language `#1'. Using the pattern for}%
\typeout{** the default language instead.}%
\else
\language=\csname l@#1\endcsname
\fi
#2}}
\providecommand{\BIBdecl}{\relax}
\BIBdecl

\bibitem{wan2014gesture}
Q.~Wan, Y.~Li, C.~Li, and R.~Pal, ``Gesture recognition for smart home
  applications using portable radar sensors,'' in \emph{2014 36th annual
  international conference of the IEEE engineering in medicine and biology
  society}.\hskip 1em plus 0.5em minus 0.4em\relax IEEE, 2014, pp. 6414--6417.

\bibitem{ohn2014hand}
E.~Ohn-Bar and M.~M. Trivedi, ``Hand gesture recognition in real time for
  automotive interfaces: A multimodal vision-based approach and evaluations,''
  \emph{IEEE transactions on intelligent transportation systems}, vol.~15,
  no.~6, pp. 2368--2377, 2014.

\bibitem{liu2018gesture}
H.~Liu and L.~Wang, ``Gesture recognition for human-robot collaboration: A
  review,'' \emph{International Journal of Industrial Ergonomics}, vol.~68, pp.
  355--367, 2018.

\bibitem{kalgaonkar2009one}
K.~Kalgaonkar and B.~Raj, ``One-handed gesture recognition using ultrasonic
  doppler sonar,'' in \emph{2009 IEEE International Conference on Acoustics,
  Speech and Signal Processing}.\hskip 1em plus 0.5em minus 0.4em\relax IEEE,
  2009, pp. 1889--1892.

\bibitem{przybyla201412}
R.~J. Przybyla, H.-Y. Tang, S.~E. Shelton, D.~A. Horsley, and B.~E. Boser,
  ``12.1 3d ultrasonic gesture recognition,'' in \emph{2014 IEEE International
  Solid-State Circuits Conference Digest of Technical Papers (ISSCC)}.\hskip
  1em plus 0.5em minus 0.4em\relax IEEE, 2014, pp. 210--211.

\bibitem{lu2014hand}
Z.~Lu, X.~Chen, Q.~Li, X.~Zhang, and P.~Zhou, ``A hand gesture recognition
  framework and wearable gesture-based interaction prototype for mobile
  devices,'' \emph{IEEE transactions on human-machine systems}, vol.~44, no.~2,
  pp. 293--299, 2014.

\bibitem{zhang2015tomo}
Y.~Zhang and C.~Harrison, ``Tomo: Wearable, low-cost electrical impedance
  tomography for hand gesture recognition,'' in \emph{Proceedings of the 28th
  Annual ACM Symposium on User Interface Software \& Technology}, 2015, pp.
  167--173.

\bibitem{gupta2016online}
P.~Gupta, K.~Kautz \emph{et~al.}, ``Online detection and classification of
  dynamic hand gestures with recurrent 3d convolutional neural networks,'' in
  \emph{CVPR}, vol.~1, no.~2, 2016, p.~3.

\bibitem{pisharady2015recent}
P.~K. Pisharady and M.~Saerbeck, ``Recent methods and databases in vision-based
  hand gesture recognition: A review,'' \emph{Computer Vision and Image
  Understanding}, vol. 141, pp. 152--165, 2015.

\bibitem{yun2009automatic}
L.~Yun and Z.~Peng, ``An automatic hand gesture recognition system based on
  viola-jones method and svms,'' in \emph{2009 Second International Workshop on
  Computer Science and Engineering}, vol.~2.\hskip 1em plus 0.5em minus
  0.4em\relax IEEE, 2009, pp. 72--76.

\bibitem{huang2009vision}
D.-Y. Huang, W.-C. Hu, and S.-H. Chang, ``Vision-based hand gesture recognition
  using pca+ gabor filters and svm,'' in \emph{2009 fifth international
  conference on intelligent information hiding and multimedia signal
  processing}.\hskip 1em plus 0.5em minus 0.4em\relax IEEE, 2009, pp. 1--4.

\bibitem{molchanov2016online}
P.~Molchanov, X.~Yang, S.~Gupta, K.~Kim, S.~Tyree, and J.~Kautz, ``Online
  detection and classification of dynamic hand gestures with recurrent 3d
  convolutional neural network,'' in \emph{Proceedings of the IEEE conference
  on computer vision and pattern recognition}, 2016, pp. 4207--4215.

\bibitem{yang2018making}
X.~Yang, P.~Molchanov, and J.~Kautz, ``Making convolutional networks recurrent
  for visual sequence learning,'' in \emph{Proceedings of the IEEE Conference
  on Computer Vision and Pattern Recognition}, 2018, pp. 6469--6478.

\bibitem{abavisani2019improving}
M.~Abavisani, H.~R.~V. Joze, and V.~M. Patel, ``Improving the performance of
  unimodal dynamic hand-gesture recognition with multimodal training,'' in
  \emph{Proceedings of the IEEE/CVF Conference on Computer Vision and Pattern
  Recognition}, 2019.

\bibitem{kim2016hand}
Y.~Kim and B.~Toomajian, ``Hand gesture recognition using micro-doppler
  signatures with convolutional neural network,'' \emph{IEEE Access}, vol.~4,
  pp. 7125--7130, 2016.

\bibitem{qi2017pointnet}
C.~R. Qi, H.~Su, K.~Mo, and L.~J. Guibas, ``Pointnet: Deep learning on point
  sets for 3d classification and segmentation,'' in \emph{Proceedings of the
  IEEE conference on computer vision and pattern recognition}, 2017, pp.
  652--660.

\bibitem{min2020pointlstm}
Y.~{Min}, Y.~{Zhang}, X.~{Chai}, and X.~{Chen}, ``An efficient pointlstm for
  point clouds based gesture recognition,'' in \emph{2020 IEEE/CVF Conference
  on Computer Vision and Pattern Recognition (CVPR)}, 2020, pp. 5760--5769.

\bibitem{palip2021pantomime}
S.~Palipana, D.~Salami, L.~A. Leiva, and S.~Sigg, ``Pantomime: Mid-air gesture
  recognition with sparse millimeter-wave radar point clouds,''
  \emph{Proceedings of the ACM on Interactive, Mobile, Wearable and Ubiquitous
  Technologies}, vol.~5, no.~1, pp. 1--27, 2021.

\bibitem{salami2020motion}
D.~Salami, S.~Palipana, M.~Kodali, and S.~Sigg, ``Motion pattern recognition in
  4d point clouds,'' in \emph{2020 IEEE 30th International Workshop on Machine
  Learning for Signal Processing (MLSP)}.\hskip 1em plus 0.5em minus
  0.4em\relax IEEE, 2020, pp. 1--6.

\bibitem{su2015multi}
H.~Su, S.~Maji, E.~Kalogerakis, and E.~Learned-Miller, ``Multi-view
  convolutional neural networks for 3d shape recognition,'' in
  \emph{Proceedings of the IEEE international conference on computer vision},
  2015, pp. 945--953.

\bibitem{yu2018multi}
T.~Yu, J.~Meng, and J.~Yuan, ``Multi-view harmonized bilinear network for 3d
  object recognition,'' in \emph{Proceedings of the IEEE Conference on Computer
  Vision and Pattern Recognition}, 2018, pp. 186--194.

\bibitem{maturana2015voxnet}
D.~Maturana and S.~Scherer, ``Voxnet: A 3d convolutional neural network for
  real-time object recognition,'' in \emph{2015 IEEE/RSJ International
  Conference on Intelligent Robots and Systems (IROS)}.\hskip 1em plus 0.5em
  minus 0.4em\relax IEEE, 2015, pp. 922--928.

\bibitem{riegler2017octnet}
G.~Riegler, A.~Osman~Ulusoy, and A.~Geiger, ``Octnet: Learning deep 3d
  representations at high resolutions,'' in \emph{Proceedings of the IEEE
  Conference on Computer Vision and Pattern Recognition}, 2017, pp. 3577--3586.

\bibitem{wang2019dynamic}
Y.~Wang, Y.~Sun, Z.~Liu, S.~E. Sarma, M.~M. Bronstein, and J.~M. Solomon,
  ``Dynamic graph cnn for learning on point clouds,'' \emph{Acm Transactions On
  Graphics (tog)}, vol.~38, no.~5, pp. 1--12, 2019.

\bibitem{odoemelem2020low}
H.~U. Odoemelem and K.~Van~Laerhoven, ``A low-cost prototyping framework for
  human-robot desk interaction,'' in \emph{Adjunct Proceedings of the 2020 ACM
  International Joint Conference on Pervasive and Ubiquitous Computing and
  Proceedings of the 2020 ACM International Symposium on Wearable Computers},
  2020, pp. 191--194.

\bibitem{soni2018artificial}
V.~D. Soni, ``Artificial cognition for human-robot interaction,''
  \emph{International Journal on Integrated Education}, vol.~1, no.~1, pp.
  49--53, 2018.

\bibitem{vujovic2015raspberry}
V.~Vujovi{\'c} and M.~Maksimovi{\'c}, ``Raspberry pi as a sensor web node for
  home automation,'' \emph{Computers \& Electrical Engineering}, vol.~44, pp.
  153--171, 2015.

\bibitem{jain2014raspberry}
S.~Jain, A.~Vaibhav, and L.~Goyal, ``Raspberry pi based interactive home
  automation system through e-mail,'' in \emph{2014 International Conference on
  Reliability Optimization and Information Technology (ICROIT)}.\hskip 1em plus
  0.5em minus 0.4em\relax IEEE, 2014, pp. 277--280.

\bibitem{Wachs11}
J.~P. Wachs, M.~Kölsch, H.~Stern, and Y.~Edan, ``Vision-based hand-gesture
  applications,'' \emph{Commun. ACM}, vol.~54, no.~2, 2011.

\bibitem{Rautaray12}
S.~S. Rautaray and A.~Agrawal, ``Vision based hand gesture recognition for
  human computer interaction: A survey.'' \emph{Artif. Intell. Rev.}, vol.~43,
  no.~1, pp. 1--54, 2012.

\bibitem{lun2015survey}
R.~Lun and W.~Zhao, ``A survey of applications and human motion recognition
  with {Microsoft} {Kinect},'' \emph{Int. J. Pattern Recognit. Artif. Intell.},
  vol.~29, no.~5, pp. 1\,555\,008:1--48, 2015.

\bibitem{Caine12_privacy}
K.~Caine, S.~\v{S}abanovic, and M.~Carter, ``The effect of monitoring by
  cameras and robots on the privacy enhancing behaviors of older adults,'' in
  \emph{Proc. HRI}, 2012, pp. 343--350.

\bibitem{abdelnasser2015wigest}
H.~Abdelnasser, M.~Youssef, and K.~A. Harras, ``{WiGest}: A ubiquitous
  wifi-based gesture recognition system,'' in \emph{Proc. INFOCOM}, 2015.

\bibitem{li2016wifinger}
H.~Li, W.~Yang, J.~Wang, Y.~Xu, and L.~Huang, ``{WiFinger}: talk to your smart
  devices with finger-grained gesture,'' in \emph{Proc. UbiComp}, 2016, pp.
  250--261.

\bibitem{ma2018signfi}
Y.~Ma, G.~Zhou, S.~Wang, H.~Zhao, and W.~Jung, ``{SignFi}: Sign language
  recognition using wifi,'' \emph{ACM IMWUT}, vol.~2, no.~1, pp. 1--21, 2018.

\bibitem{venkatnarayan2018multi}
R.~H. Venkatnarayan, G.~Page, and M.~Shahzad, ``Multi-user gesture recognition
  using {WiFi},'' in \emph{Proc. MobiSys}, 2018, pp. 401--413.

\bibitem{virmani2017position}
A.~Virmani and M.~Shahzad, ``Position and orientation agnostic gesture
  recognition using {WiFi},'' in \emph{Proc. MobiSys}, 2017, pp. 252--264.

\bibitem{pu2013whole}
Q.~Pu, S.~Gupta, S.~Gollakota, and S.~Patel, ``Whole-home gesture recognition
  using wireless signals,'' in \emph{Proc. MobiCom}, 2013, pp. 27--38.

\bibitem{li2019making}
T.~Li, L.~Fan, M.~Zhao, Y.~Liu, and D.~Katabi, ``Making the invisible visible:
  Action recognition through walls and occlusions,'' in \emph{Proc. ICCV},
  2019, pp. 872--881.

\bibitem{zhao2019through}
M.~Zhao, Y.~Liu, A.~Raghu, T.~Li, H.~Zhao, A.~Torralba, and D.~Katabi,
  ``Through-wall human mesh recovery using radio signals,'' in \emph{Proc.
  ICCV}, 2019, pp. 10\,113--10\,122.

\bibitem{lien2016soli}
J.~Lien, N.~Gillian, M.~E. Karagozler, P.~Amihood, C.~Schwesig, E.~Olson,
  H.~Raja, and I.~Poupyrev, ``Soli: Ubiquitous gesture sensing with millimeter
  wave radar,'' \emph{ACM Trans. Graphics}, vol.~35, no.~4, pp. 1--19, 2016.

\bibitem{wei2015mtrack}
T.~Wei and X.~Zhang, ``{mTrack}: High-precision passive tracking using
  millimeter wave radios,'' in \emph{Proc. MobiCom}, 2015, pp. 117--129.

\bibitem{Berenguer2019GestureVLAD}
A.~D. Berenguer, M.~C. Oveneke, H.~Khalid, M.~Alioscha-Perez, A.~Bourdoux, and
  H.~Sahli, ``{GestureVLAD}: Combining unsupervised features representation and
  spatio-temporal aggregation for doppler-radar gesture recognition,''
  \emph{IEEE Access}, vol.~7, pp. 137\,122--137\,135, 2019.

\bibitem{panneer2020mmASL}
P.~S. Santhalingam, A.~A. Hosain, D.~Zhang, P.~Pathak, H.~Rangwala, and
  R.~Kushalnagar, ``{Environment-Independent {ASL} Gesture Recognition Using
  {60 GHz} Millimeter-wave Signals},'' \emph{ACM IMWUT}, vol.~4, no.~1, pp.
  1--30, 2020.

\bibitem{singh2019radhar}
A.~D. Singh, S.~S. Sandha, L.~Garcia, and M.~Srivastava, ``Radhar: Human
  activity recognition from point clouds generated through a millimeter-wave
  radar,'' in \emph{Proceedings of the 3rd ACM Workshop on Millimeter-wave
  Networks and Sensing Systems}, 2019, pp. 51--56.

\bibitem{wang2016interacting}
S.~Wang, J.~Song, J.~Lien, I.~Poupyrev, and O.~Hilliges, ``Interacting with
  {Soli}: Exploring fine-grained dynamic gesture recognition in the
  radio-frequency spectrum,'' in \emph{Proc. UIST}, 2016, pp. 851--860.

\bibitem{zhao2019mid}
P.~Zhao, C.~X. Lu, J.~Wang, C.~Chen, W.~Wang, N.~Trigoni, and A.~Markham,
  ``mid: Tracking and identifying people with millimeter wave radar,'' in
  \emph{In Proc. of DCOSS}.\hskip 1em plus 0.5em minus 0.4em\relax IEEE, 2019,
  pp. 33--40.

\bibitem{meng2020gait}
Z.~Meng, S.~Fu, J.~Yan, H.~Liang, A.~Zhou, S.~Zhu, H.~Ma, J.~Liu, and N.~Yang,
  ``Gait recognition for co-existing multiple people using millimeter wave
  sensing,'' in \emph{In Proc. of AAAI}, vol.~34, no.~01, 2020, pp. 849--856.

\bibitem{qian20203d}
K.~Qian, Z.~He, and X.~Zhang, ``3d point cloud generation with millimeter-wave
  radar,'' \emph{Proceedings of the ACM on Interactive, Mobile, Wearable and
  Ubiquitous Technologies}, vol.~4, no.~4, pp. 1--23, 2020.

\bibitem{dong2020model}
Z.~Dong, F.~Li, J.~Ying, and K.~Pahlavan, ``A model-based rf hand motion
  detection system for shadowing scenarios,'' \emph{IEEE Access}, vol.~8, pp.
  115\,662--115\,672, 2020.

\bibitem{liu2020real}
H.~Liu, Y.~Wang, A.~Zhou, H.~He, W.~Wang, K.~Wang, P.~Pan, Y.~Lu, L.~Liu, and
  H.~Ma, ``Real-time arm gesture recognition in smart home scenarios via
  millimeter wave sensing,'' \emph{Proceedings of the ACM on Interactive,
  Mobile, Wearable and Ubiquitous Technologies}, vol.~4, no.~4, pp. 1--28,
  2020.

\bibitem{qi2017pointnet2}
C.~R. Qi, L.~Yi, H.~Su, and L.~J. Guibas, ``Pointnet++: Deep hierarchical
  feature learning on point sets in a metric space,'' in \emph{Proceedings of
  the 31st International Conference on Neural Information Processing Systems},
  ser. NIPS'17, 2017.

\bibitem{Kipf:2016tc}
T.~N. Kipf and M.~Welling, ``{Semi-Supervised Classification with Graph
  Convolutional Networks},'' in \emph{Proceedings of the 5th International
  Conference on Learning Representations}, ser. ICLR '17, 2017.

\bibitem{2017Gimessagepassing}
J.~Gilmer, S.~S. Schoenholz, P.~F. Riley, O.~Vinyals, and G.~E. Dahl, ``Neural
  message passing for quantum chemistry,'' in \emph{Proceedings of the 34th
  International Conference on Machine Learning - Volume 70}, ser. ICML'17,
  2017.

\bibitem{wang2019edgeconv}
\BIBentryALTinterwordspacing
Y.~Wang, Y.~Sun, Z.~Liu, S.~E. Sarma, M.~M. Bronstein, and J.~M. Solomon,
  ``Dynamic graph cnn for learning on point clouds,'' \emph{ACM Trans. Graph.},
  vol.~38, no.~5, Oct. 2019. [Online]. Available:
  \url{https://doi.org/10.1145/3326362}
\BIBentrySTDinterwordspacing

\bibitem{owoyemi2018spatiotemporal}
J.~Owoyemi and K.~Hashimoto, ``Spatiotemporal learning of dynamic gestures from
  3d point cloud data,'' in \emph{2018 IEEE International Conference on
  Robotics and Automation (ICRA)}.\hskip 1em plus 0.5em minus 0.4em\relax IEEE,
  2018, pp. 1--5.

\bibitem{vaswani2017attention}
A.~Vaswani, N.~Shazeer, N.~Parmar, J.~Uszkoreit, L.~Jones, A.~N. Gomez,
  L.~Kaiser, and I.~Polosukhin, ``Attention is all you need,'' \emph{arXiv
  preprint arXiv:1706.03762}, 2017.

\bibitem{richards2010principles}
M.~A. Richards, J.~A. Scheer, and W.~A. Holm, \emph{Principles of Modern Radar:
  Basic Principles}.\hskip 1em plus 0.5em minus 0.4em\relax Scitech Publishing,
  2010.

\bibitem{jaderberg2015transform}
M.~Jaderberg, K.~Simonyan, A.~Zisserman, and k.~kavukcuoglu, ``Spatial
  transformer networks,'' in \emph{Advances in Neural Information Processing
  Systems}, vol.~28.\hskip 1em plus 0.5em minus 0.4em\relax Curran Associates,
  Inc., 2015, pp. 2017--2025.

\bibitem{vas2017attention}
A.~Vaswani, N.~Shazeer, N.~Parmar, J.~Uszkoreit, L.~Jones, A.~N. Gomez, L.~u.
  Kaiser, and I.~Polosukhin, ``Attention is all you need,'' in \emph{Advances
  in Neural Information Processing Systems}, I.~Guyon, U.~V. Luxburg,
  S.~Bengio, H.~Wallach, R.~Fergus, S.~Vishwanathan, and R.~Garnett, Eds.,
  vol.~30.\hskip 1em plus 0.5em minus 0.4em\relax Curran Associates, Inc.,
  2017, pp. 5998--6008.

\bibitem{cohen2006learning}
G.~Cohen, M.~Hilario, H.~Sax, S.~Hugonnet, and A.~Geissbuhler, ``Learning from
  imbalanced data in surveillance of nosocomial infection,'' \emph{Artificial
  intelligence in medicine}, vol.~37, no.~1, pp. 7--18, 2006.

\bibitem{paszke2019pytorch}
A.~Paszke, S.~Gross, F.~Massa, A.~Lerer, J.~Bradbury, G.~Chanan, T.~Killeen,
  Z.~Lin, N.~Gimelshein, L.~Antiga \emph{et~al.}, ``Pytorch: An imperative
  style, high-performance deep learning library,'' \emph{arXiv preprint
  arXiv:1912.01703}, 2019.

\bibitem{fey2019fast}
M.~Fey and J.~E. Lenssen, ``Fast graph representation learning with pytorch
  geometric,'' \emph{arXiv preprint arXiv:1903.02428}, 2019.

\bibitem{kingma2014adam}
D.~P. Kingma and J.~Ba, ``Adam: A method for stochastic optimization,''
  \emph{arXiv preprint arXiv:1412.6980}, 2014.

\bibitem{ruiz2011doubleflip}
J.~Ruiz and Y.~Li, ``Doubleflip: a motion gesture delimiter for mobile
  interaction,'' in \emph{Proceedings of the SIGCHI Conference on Human Factors
  in Computing Systems}, 2011, pp. 2717--2720.

\bibitem{kerber2015wristrotate}
F.~Kerber, P.~Schardt, and M.~L{\"o}chtefeld, ``Wristrotate: a personalized
  motion gesture delimiter for wrist-worn devices,'' in \emph{Proceedings of
  the 14th international conference on mobile and ubiquitous multimedia}, 2015,
  pp. 218--222.

\bibitem{masina2020investigating}
F.~Masina, V.~Orso, P.~Pluchino, G.~Dainese, S.~Volpato, C.~Nelini, D.~Mapelli,
  A.~Spagnolli, and L.~Gamberini, ``Investigating the accessibility of voice
  assistants with impaired users: Mixed methods study,'' \emph{Journal of
  medical Internet research}, vol.~22, no.~9, p. e18431, 2020.

\bibitem{de20173d}
Q.~De~Smedt, H.~Wannous, J.-P. Vandeborre, J.~Guerry, B.~L. Saux, and
  D.~Filliat, ``3d hand gesture recognition using a depth and skeletal dataset:
  Shrec'17 track,'' in \emph{Proceedings of the Workshop on 3D Object
  Retrieval}, 2017, pp. 33--38.

\end{thebibliography}

\begin{IEEEbiography}[{\includegraphics[width=1in,height=1.25in,clip,keepaspectratio]{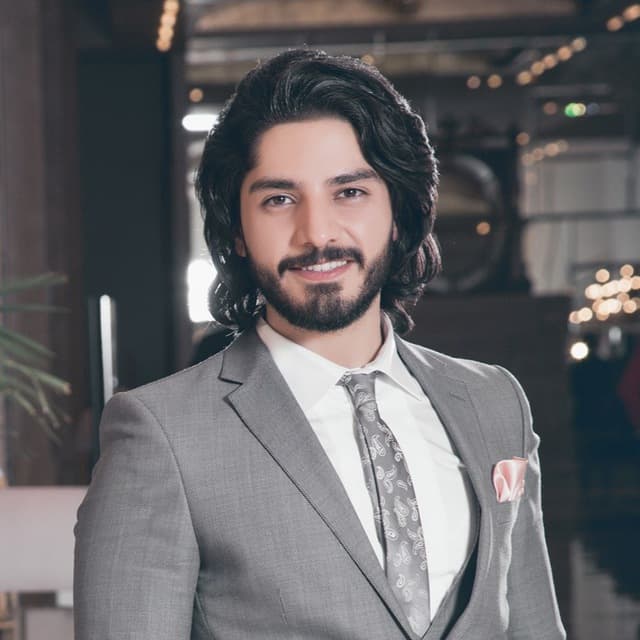}}]{Dariush Salami}
received his BSc and MSc degrees from Shahid Beheshti University and Amirkabir University of Technology in Software Engineering in 2016 and 2019, respectively. He is currently a Marie Skłodowska Curie fellow in ITN-WindMill project and a PhD researcher at the department of communications and networking at Aalto University. He is mainly focused on Machine Learning for Wireless Communications and Sensing especially in mmWave range.
\end{IEEEbiography}

\begin{IEEEbiography}[{\includegraphics[width=1in,height=1.25in,clip,keepaspectratio]{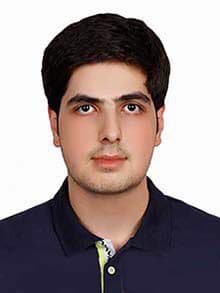}}]{Ramin Hasibi}
received his the BSc and MSc from Isfahan University of Technology and Amirkabir University of Technology in Information Technology Engineering in 2016 and 2019, respectively. He is currently a Ph.D. researcher at the department of informatics, University of Bergen where his main research focus is on Graph Representation Learning and Graph Neural Networks as well as their application in different domains.
\end{IEEEbiography}

\begin{IEEEbiography}[{\includegraphics[width=1in,height=1.25in,clip,keepaspectratio]{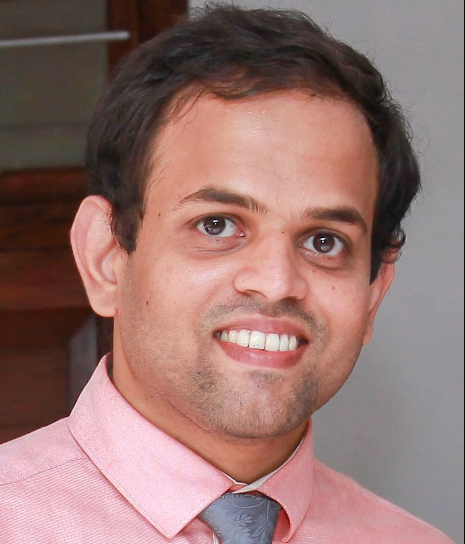}}]{Sameera Palipana}
is currently a System Specification Engineer at Nokia Solutions and Networks, Espoo and a Visiting Researcher at Aalto University. He was a Postdoctoral Researcher at Aalto University between 2019-2021. He  obtained his PhD from Munster Technological University, Ireland, in 2019, received his Master's degree at University of Bremen, Germany in 2014 in Information and Communication Technology, and received his B.Sc. (Hons) degree in Electronics and Telecommunication Engineering from University of Moratuwa, Sri Lanka in 2010.  
\end{IEEEbiography}

\begin{IEEEbiography}[{\includegraphics[width=1in,height=1.25in,clip,keepaspectratio]{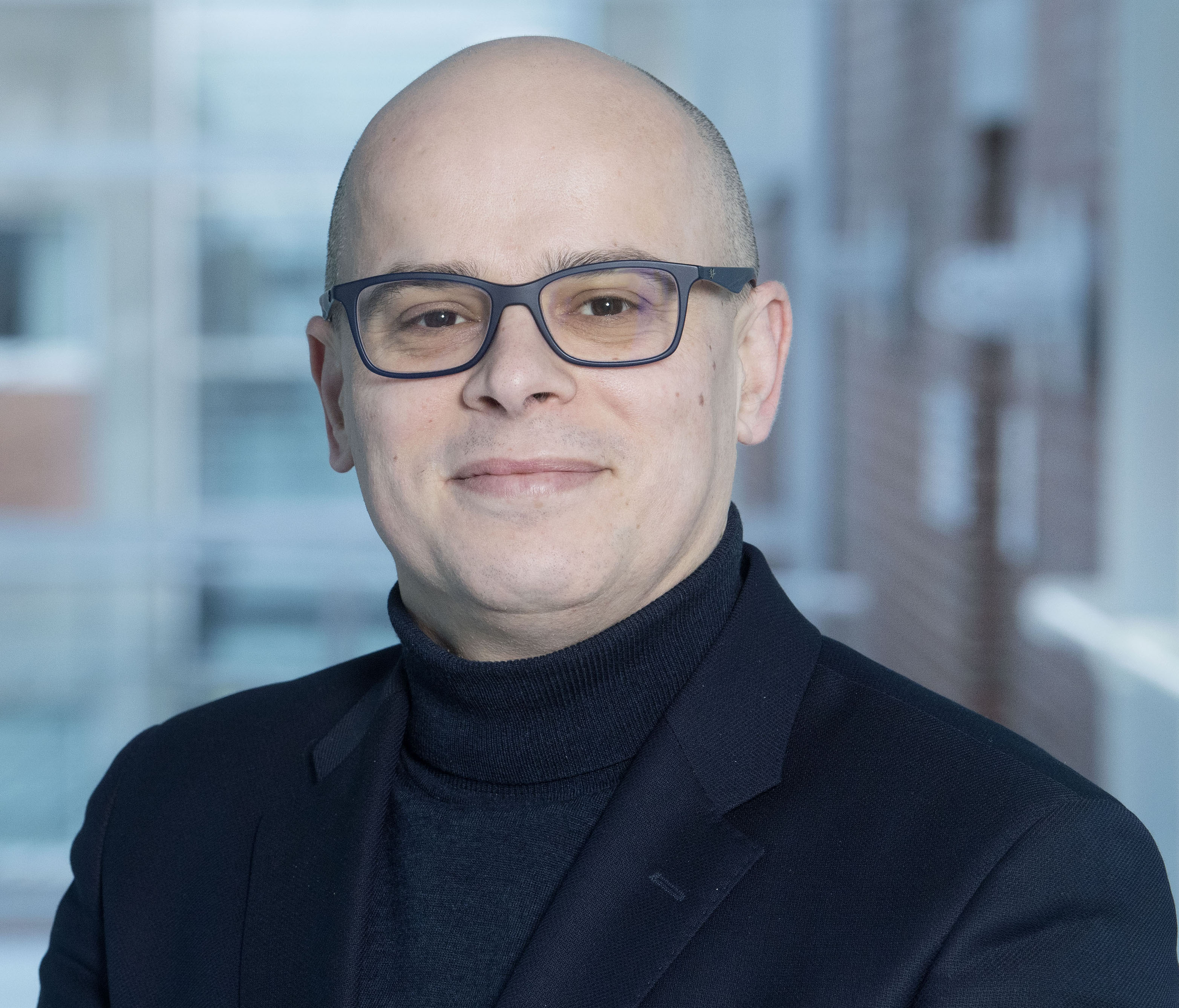}}]{Petar Popovski}
is a Professor at Aalborg University, where he heads the section on Connectivity and a Visiting Excellence Chair at the University of Bremen. He received his Dipl.-Ing and M. Sc. degrees in  communication  engineering  from  the  University  of  Sts.  Cyril  and  Methodius  in  Skopje  and  the Ph.D.  degree  from  Aalborg  University  in  2005.  He  is  a  Fellow  of  the  IEEE.  He  received  an  ERC Consolidator  Grant  (2015),  the  Danish  Elite  Researcher  award  (2016),  IEEE  Fred  W.  Ellersick  prize (2016),  IEEE  Stephen  O.  Rice  prize  (2018),  Technical  Achievement  Award  from  the  IEEE  Technical Committee on Smart Grid Communications (2019), the Danish Telecommunication Prize (2020) and Villum  Investigator  Grant  (2021).  He  is  a  Member  at  Large  at  the  Board  of  Governors  in  IEEE Communication  Society,  Vice-Chair  of  the  IEEE  Communication  Theory  Technical  Committee  and IEEE  TRANSACTIONS  ON  GREEN  COMMUNICATIONS  AND  NETWORKING.  He  is  currently  an  Area Editor of the IEEE TRANSACTIONS ON WIRELESS COMMUNICATIONS. Prof. Popovski was the General Chair for IEEE SmartGridComm 2018 and IEEE Communication Theory Workshop 2019. His research interests  are  in  the  area  of  wireless  communication  and  communication  theory.  He  authored  the
book ``Wireless Connectivity: An Intuitive and Fundamental Guide'', published by Wiley in 2020.
\end{IEEEbiography}

\begin{IEEEbiography}[{\includegraphics[width=1in,height=1.25in,clip,keepaspectratio]{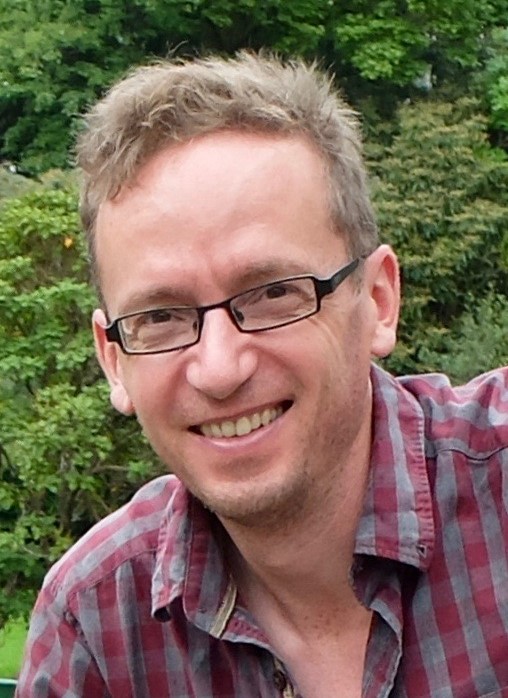}}]{Tom Michoel}
is Professor in bioinformatics at the Computational Biology Unit at the Department of Informatics at the University of Bergen since 2018, and was an independent group leader in computational biology at the University of Edinburgh (2012-2018) and the University of Freiburg (2010-2012). He obtained the MSc degree in Physics (1997) and PhD degree in Mathematical Physics (2001) from the KU Leuven, and was a postdoctoral researcher in mathematics (UC Davis, 2001-2002), theoretical physics (KU Leuven, 2002-2004), and bioinformatics and systems biology (Ghent University, 2004-2010). His research focus in the last five years has been on developing methods, algorithms, and software for causal inference and Bayesian network learning from high-dimensional omics data, supported by grants from the BBSRC (2015-2016), the NIH (2016-2019), the MRC (2017-2021), and the Norwegian Research Council (2021-2024).
\end{IEEEbiography}

\begin{IEEEbiography}[{\includegraphics[width=1in,height=1.25in,clip,keepaspectratio]{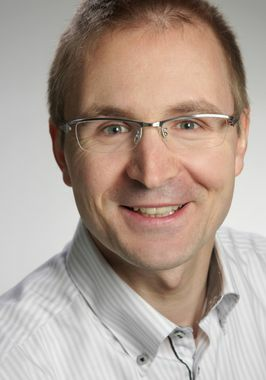}}]{Stephan Sigg}
 received his M.Sc. degree in computer science from TU Dortmund,
in 2004 and his Ph.D. degree from Kassel University, in 2008. Since 2015 he is
an assistant professor at Aalto University, Finland. He is a member of the
editorial board of the Proceedings of the ACM on Interactive, Mobile, Wearable
and Ubiquitous Technologies as well as of the Elsevier journal of Computer
Communications. He has served as a TPC member of renowned conferences
including IEEE PerCom, IEEE ICDCS, etc. His research interests include Ambient
Intelligence, in particular, Pervasive sensing, activity recognition, usable
security algorithms for mobile distributed systems.
\end{IEEEbiography}
\end{document}